\documentclass{article}
\pdfoutput=1

\usepackage[dvipsnames,table]{xcolor}

\definecolor{qual-fig-green}{RGB}{0,144,11}
\definecolor{qual-fig-red}{RGB}{238,0,0}
\definecolor{qual-fig-purple}{RGB}{153,51,255}
\definecolor{cvprblue}{rgb}{0.21,0.49,0.74}
\definecolor{citecolor}{HTML}{0071BC}
\definecolor{linkcolor}{HTML}{ED1C24}

\usepackage[preprint,nonatbib]{neurips_2023}

\usepackage[utf8]{inputenc}
\usepackage[T1]{fontenc}
\usepackage{url}
\usepackage{booktabs}
\usepackage{amsfonts}
\usepackage{amsmath}
\usepackage{nicefrac}
\usepackage{microtype}
\usepackage{lipsum}
\usepackage{graphicx}
\usepackage{wrapfig}
\usepackage[size=small]{caption}
\usepackage{mathtools}
\usepackage{amssymb}
\usepackage{amsthm}
\usepackage[algoruled,boxed,lined,noend]{algorithm2e}
\usepackage{adjustbox}
\usepackage{float}
\usepackage{pifont}
\usepackage{subcaption}
\usepackage{tcolorbox}
\usepackage{pgfgantt}
\usepackage{multirow}
\usepackage[superscript,biblabel]{cite}

\usepackage{hyperref}

\usepackage[capitalize]{cleveref}

\title{Best Practices for Large Language Models in Radiology}
\author{\normalfont
Christian Bluethgen$^{1,2,\ddagger}$ 
\quad Dave Van Veen$^{1,3}$ 
\quad Cyril Zakka$^{4,5}$ \\
Katherine E Link$^{6,7}$ 
\quad Aaron Hunter Fanous$^{8}$  
\quad Roxana Daneshjou$^{9,10}$ \\
\quad Thomas Frauenfelder$^{2}$ 
\quad Curtis Langlotz$^{1,10,11,12}$ \\
\quad Sergios Gatidis$^{1,12}$ 
\quad Akshay Chaudhari$^{1,10,12}$ \\
\vspace{3mm}
\\
$^1$Stanford Center for Artificial Intelligence in Medicine and Imaging, Palo Alto, CA, USA \\
$^2$Diagnostic and Interventional Radiology, University Hospital Zurich, University of Zurich, Zurich, Switzerland \\
$^3$Department of Electrical Engineering, Stanford University, Stanford, CA, USA \\
$^4$Department of Cardiothoracic Surgery, Stanford Medicine, Stanford, CA, USA \\
$^5$Hugging Face, Manhattan, New York City, NY, USA \\
$^6$Department of Medical Education, Icahn School of Medicine at Mount Sinai, New York, NY, USA \\
$^7$NVIDIA Corporation, New York City, NY, USA \\
$^8$UT Health San Antonio, San Antonio, TX, USA \\
$^9$Department of Dermatology, Redwood City, CA, USA \\
$^{10}$Department of Biomedical Data Science, Stanford, CA, USA \\
$^{11}$Department of Medicine, Stanford, CA, USA \\
$^{12}$Department of Radiology, Stanford University, Stanford, CA, USA
}
\begin{document}
\maketitle
\begin{abstract}
At the heart of radiological practice is the challenge of integrating complex imaging data with clinical information to produce actionable insights. Nuanced application of language is key for various activities, including managing requests, describing and interpreting imaging findings in the context of clinical data, and concisely documenting and communicating the outcomes. The emergence of large language models (LLMs) offers an opportunity to improve the management and interpretation of the vast data in radiology. Despite being primarily general-purpose, these advanced computational models demonstrate impressive capabilities in specialized language-related tasks, even without specific training. Unlocking the potential of LLMs for radiology requires basic understanding of their foundations and a strategic approach to navigate their idiosyncrasies. This review, drawing from practical radiology and machine learning expertise and recent literature, provides readers insight into the potential of LLMs in radiology. It examines best practices that have so far stood the test of time in the rapidly evolving landscape of LLMs. This includes practical advice for optimizing LLM characteristics for radiology practices along with limitations, effective prompting, and fine-tuning strategies.
\end{abstract}
\footnotetext{$^\ddagger$Corresponding author.}

\newpage

\section{Introduction}
\label{sec:intro}

\begin{quote}
``All right,'' said Deep Thought. ``The Answer to the Great Question\ldots{}''\\
``Yes\ldots{}!''\\
``Of Life, the Universe and Everything\ldots{}'' said Deep Thought.\\
``Yes\ldots{}!''\\
``Is\ldots{}'' said Deep Thought, and paused.\\
``Yes\ldots{}!''\\
``Is\ldots{}''\\
``Yes\ldots{}!!!\ldots{}?''\\
``Forty-two,'' said Deep Thought, with infinite majesty and calm.''\\
\begin{flushright}
    --- Douglas Adams, \textit{The Hitchhiker's Guide to the Galaxy}
\end{flushright}
\end{quote}

Radiology has a key role in healthcare, bridging complex medical imaging and clinical practice. The field is tasked with ensuring the correct application of imaging modalities, describing and integrating imaging findings with the patient's history, interpreting the gathered information, and appropriately communicating and documenting the results\cite{langlotzRadiologyReportGuide2015a}. This process relies heavily on a precise use of language.

Given the recent breakthroughs in large language models (LLMs) capable of processing and generating human language, there exists a large promise for their use with the abundant textual data in radiology. While powerful general-purpose\cite{eloundouGPTsAreGPTs2023a} computational models like ChatGPT excel in many (domain-)specific tasks without additional modifications, their flexibility and adaptability comes with a degree of unpredictability, making LLMs akin to double-edged swords\cite{shenRadiologyLLMDoubleEdged2023}. Not unlike trying to obtain a sensible answer from Douglas Adams' fictional supercomputer \textit{Deep Thought}\cite{adams1979hitchhiker}, effectively interacting with LLMs and unlocking their full potential for radiology requires understanding what to expect of them, to understand their foundations, and systematic approaches for handling the new class of models' idiosyncrasies.

As the field is heavily working on identifying use cases and testing LLMs for various purposes\cite{bhayanaChatbotsLargeLanguage2024}, it is necessary to consider best practices for effective development and integration of LLMs in radiology.

\section{Technical Foundations of Language Models}

\subsection{Traditional Language Models}
Language models (LMs) are a class of probabilistic models designed to learn statistical patterns in natural language. These models predict the probability of a subsequent text part (termed token, e.g. a word or a character) based on preceding tokens in a sequence. The advent of deep learning in natural language processing (NLP)\cite{bengio2000neural} entailed training LMs using vast, web-scale amounts of text, relying on self-supervised learning, which utilizes the inherent structure of the data itself instead of additional labels. Such \textit{pretrained} LMs outperform traditional methods on a variety of practical tasks such as translation and text classification\cite{yu2019review}.

\subsection{Transformer-based LMs}
As of 2024, most modern LMs employ the transformer architecture\cite{vaswaniAttentionAllYou2017}, which first breaks up inputs like text sequences into smaller units (tokens). Transformers integrate multiple essential components, most notably 1) embeddings (numerical vector representations that encode semantic information of tokens), 2) the attention mechanism that can model relationships between input elements even over long distances in the sequence, and 3) feed-forward networks that perform further computation over these representations. The attention mechanism is particulary notable as it allows the model to dynamically weight the importance of different input parts based on their relevance in the context.

Transformer-based LMs include an encoder (which converts text into embeddings), a decoder (which converts embeddings into text), or a combination of both. For instance, BERT (Bidirectional Encoder Representations from Transformers), an encoder-only model, excels at tasks like text classification\cite{devlin2018bert}. In contrast, GPT (generative pretrained transformer), a decoder-only model, is adept at generating text\cite{brownLanguageModelsAre2020}. Tasks like translation benefit from both an encoder that processes the source text to generate meaningful, contextualized representations, and a decoder that generates the target text in another language.

\subsection{Emergence of LLMs}
The performance of transformers scales as a function of the dataset size (often measured in trillion tokens) and model size (often measured in billion parameters)\cite{kaplanScalingLawsNeural2020}, giving rise to \textit{large} language models (LLMs). The scale of LLMs grants them unprecedented versatility across a broad spectrum of tasks\cite{brownLanguageModelsAre2020}. LLMs are characterized by abilities like following natural language instructions, and in-context learning, the ability to flexibly adhere to specific tasks by providing examples in natural language at inference time (i.e., when generating text, after training). Notably, LLMs acquire these abilities even without direct, task-specific supervision\cite{brownLanguageModelsAre2020}. While LLMs are versatile, adapting them for specific tasks can further enhance their effectiveness\cite{van-veen-etal-2023-radadapt}.

\begin{figure}
    \centering
    \includegraphics[width=1.0\linewidth]{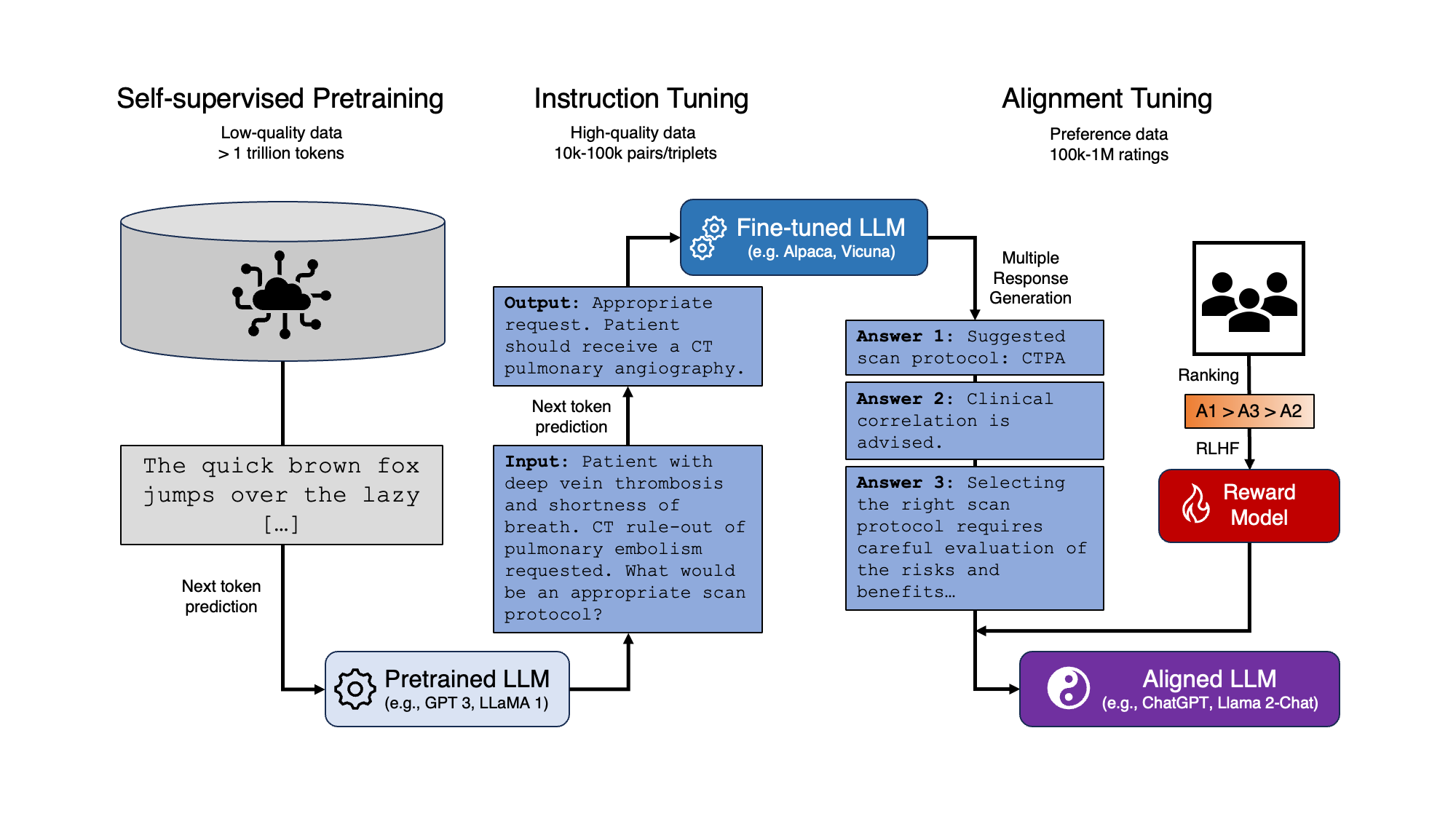}
    \caption{Process of training a large language model (LLM). (1) First, an LLM is pretrained in a self-supervised way to predict the next text parts (tokens), using vast text corpora derived from diverse, mostly web-based and weakly or uncurated sources. The resulting pretrained LLM contains general language knowledge. (2) The next step (instruction tuning) takes the pretrained model and fine-tunes it in a supervised way to predict the next tokens in input-output pairs of higher-quality data (such as curated question and answer datasets). The fine-tuned LLM has now learned to follow instructions and can solve a great variety of language tasks but can produce unexpected or potentially harmful outputs. (3) To improve this, the third step (alignment tuning) seeks to align the fine-tuned LLM with human preferences. One way to achieve this is by having the fine-tuned LLM output several answers to per input prompt. Trained human raters rank the outputs by quality. This ranking data is used to train a reward model through reinforcement learning from human feedback (RLHF). The reward model learns to predict human preference of generated outputs and is used to inform further training to create an aligned LLM. In this example, human raters preferred short, concise and specific answers (Answer 1) over more nuanced, lengthy answers (Answer 3) or answers that did not contribute to answering the question (Answer 2). CTPA: CT pulmonary angiography.}
    \label{fig:1_llm_training}
\end{figure}

Most recent LLMs are pretrained by learning to predict the most likely next token, given a sequence of input tokens\cite{yang2019xlnet}(\textbf{Fig. \ref{fig:1_llm_training}}). At this stage, the LLMs possess a general ``understanding'' of language but cannot yet respond to instructions. Therefore, the next stage aims at supervised training on high-quality data consisting of instructions and answers (instruction tuning). After this, modern LLMs can follow natural language instructions but are not yet aligned with human preferences and may respond in undesirable or even harmful ways. This necessitates methods for alignment tuning, such as reinforcement learning with human feedback (RLHF)\cite{ouyangTrainingLanguageModels2022}.

\subsection{Enhancing LLMs with Non-Textual Inputs, Retrieval-augmented Generation and Tools}
An emerging practice has seen the expansion of LLMs beyond common language tasks, evolving them into even more versatile multi-functional tools. This progression is marked by the integration of multimodal capabilities, enabling LLMs to process and interpret non-textual (e.g., visual or auditory) information. One way to achieve this is to convert images into a series of patches, akin to textual tokens, and feed them into the model alongside textual data (often after "translating" the visual tokens with the help of additional model components), enabling the output of text grounded in visual information\cite{chenVisualGPT2022,liu2023visual} - a scenario very relevant to radiological image interpretation and report writing.

\begin{figure}
    \centering
    \includegraphics[width=1.0\linewidth]{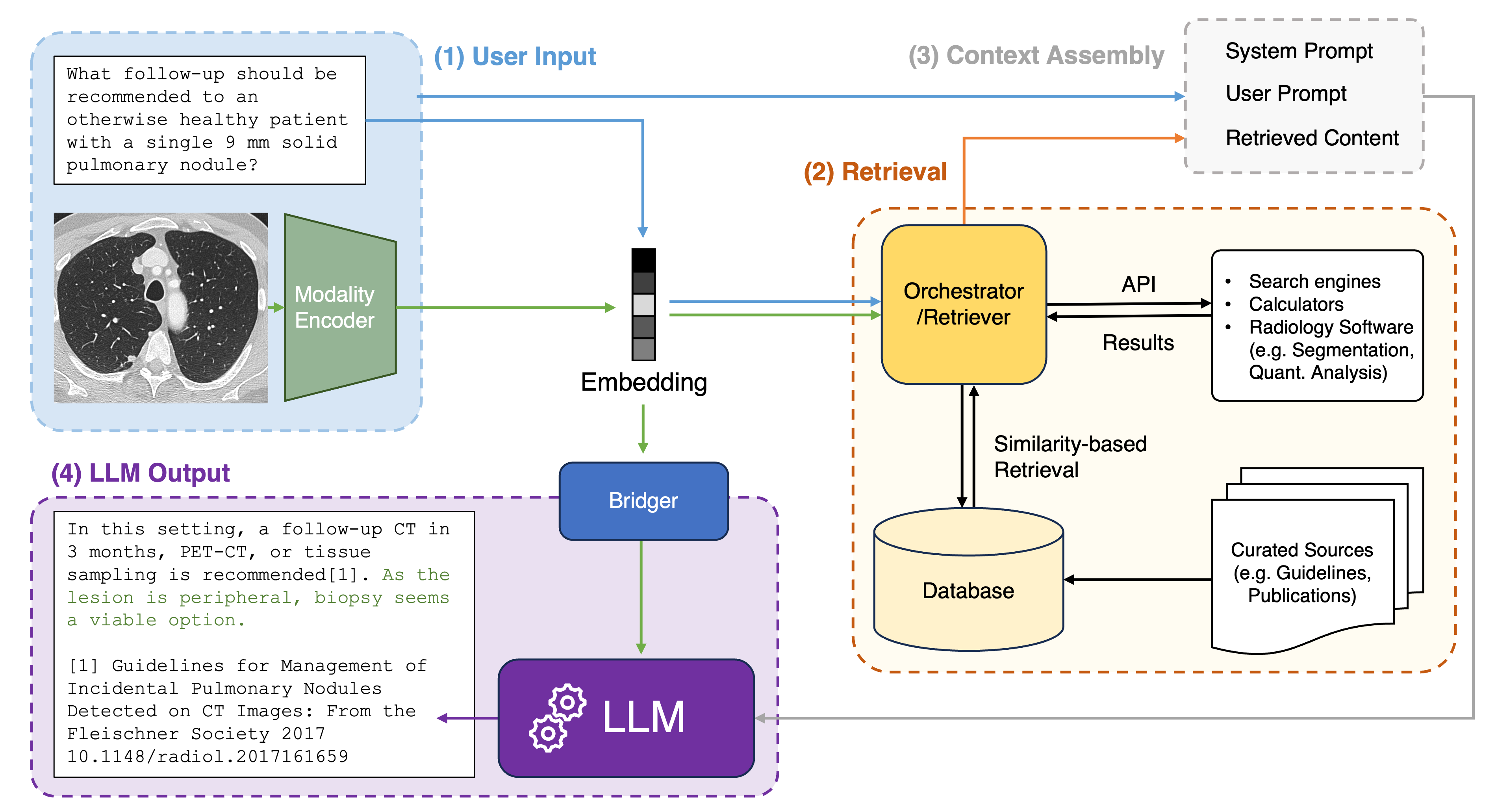}
    \caption{Simplified example of enhancing model responses by incorporating additional information from external sources with a large language model (LLM) accepting multi-modal inputs. (1) A user inputs a query and (optional) image data. This data is transformed into numerical vector representations (embeddings). (2) An orchestrator module retrieves content similar to the user query (or the image input) from an optimized database. This database can contain curated, verified high-quality information such as guidelines or scientific publications. Additionally, the orchestrator module can interact with selected software tools to retrieve additional information (e.g. from search engines, specialized radiology applications like segmentation tools, or automated analysis and quantification software). (3) The retrieved content (usually text) is concatenated with a system prompt carrying general instructions for the model and the user input with the specific request. (4) The user input and the retrieved context-specific content are provided to the LLM to generate an output. In case an image was initially provided, a bridger module (another model that can bridge the modality gap by converting visual embeddings into embeddings that the text-based LLMs can handle) can "translate" visual information for the LLM. API: Application Programming Interface.}
    \label{fig:2_llm_training}
\end{figure}

To further enhance the versatility of LLMs, additional external information can be retrieved and provided to LLMs, a process known as retrieval-augmented generation (RAG)(\textbf{Fig.~\ref{fig:2_llm_training}}). This information can come from curated text sources (e.g., databases with guidelines), or provided through the addition of external tools like browsers, calculators, and Application Programming Interfaces (APIs). This has been pivotal in enabling LLMs to access real-time information, perform complex calculations, and interact with various web services to improve the reliability and relevance of their outputs\cite{schickToolformerLanguageModels2023,nakanoWebGPTBrowserassistedQuestionanswering2022}.

\newpage
\section{LLMs in Radiology: Applications and Desiderata}

Radiology requires the use of language to acquire and interpret imaging data and to communicate key findings. Clinical context is a key for medical decision making, and its absence has been identified as a significant limitation of current medical AI systems\cite{bradyDevelopingPurchasingImplementing2024}. Through their versatility to process complex, contextualized information and to interact in natural language, LLMs are a big step towards overcoming this limitation and uniquely position them to augment workflows and substantially influence how the field of radiology operates.

\subsection{Applications}

\textbf{Clinical tasks}. Initial work has demonstrated the use of LLMs for report summarization, simplification, structuring, quality assurance, detection of speech recognition errors, as well as generating a report from structured knowledge following a preferred reporting style\cite{kimLargeLanguageModels2024,yanStyleAwareRadiologyReport2023,schmidtGeneratingLargeLanguage2024}. LLMs could potentially draft recommendations that follow from radiology reports (e.g. by referencing guidelines) or suggest actions based on those recommendations through alert and scheduling systems. Using an LLM as a copilot informed by image and text data or additional specialized software could significantly speed up report generation. For instance, preparing multidisciplinary boards is a tedious task that an LLM could support by pre-structuring case information according to the board's preferences and augment the presentation by providing relevant information derived from guidelines and scientific literature.

\textbf{Operational tasks}. LLMs could potentially increase operational efficiency and accessibility of radiology services. Exam, service and staff scheduling, reporting and resource allocation are complex tasks that require integration of information from different sources, including patient records, availability of resources and departmental guidelines. Despite standardization efforts, this information is often unstructured and distributed across several systems and could greatly benefit from enhanced general language processing abilities.

\textbf{Research}. Clinical research relies on the identification of suitable patient cohorts for conducting research studies and often requires manually sifting vast amounts of health records. This process can be facilitated by LLM-based screening of records based on defined inclusion criteria and endpoints. LLMs can also be an effective tool for retrieving and summarizing scientific literature on topics of interest to generate hypotheses\cite{tangEvaluatingLargeLanguage2023}. Increasingly, LLMs are being explicitly supported\cite{kollerWhyWeSupport2023} to assist in the communication of scientific results within traditional media (e.g. through manuscript preparation and review) or within the growing field of alternative media, such as online platforms.

\textbf{Education}. Traditionally, effective radiology training includes frequent case discussions and feedback on the reports drafted by residents by more experienced radiologists. Conversational LLMs drawing information from high-quality case files and guidelines could provide a new tool to engage with trainees and guide them towards proficiency in an interactive and individualized way.

\subsection{Desiderata for LLMs in Radiology}
As LLMs demonstrate increasing feasibility in radiological applications, the field must define clear requirements and desired properties for their effective integration into research and clinical practice. Key desiderata include: improving patient outcomes, enhancing radiologist workflow efficiency (without increasing the risks of burnout), and maintaining high standards of clinical care and privacy preservation. Pursuing these overarching desired goals requires figuring out how to use LLMs to their fullest potential, while respecting ethical principles guiding how to promote wellbeing, minimize harm, provide appropriate transparency and add value while being dependable, acting responsibly, ecologically and economically viable\cite{bradyDevelopingPurchasingImplementing2024,geisEthicsArtificialIntelligence2019}. 

\begin{figure}
    \centering
    \includegraphics[width=1.0\linewidth]{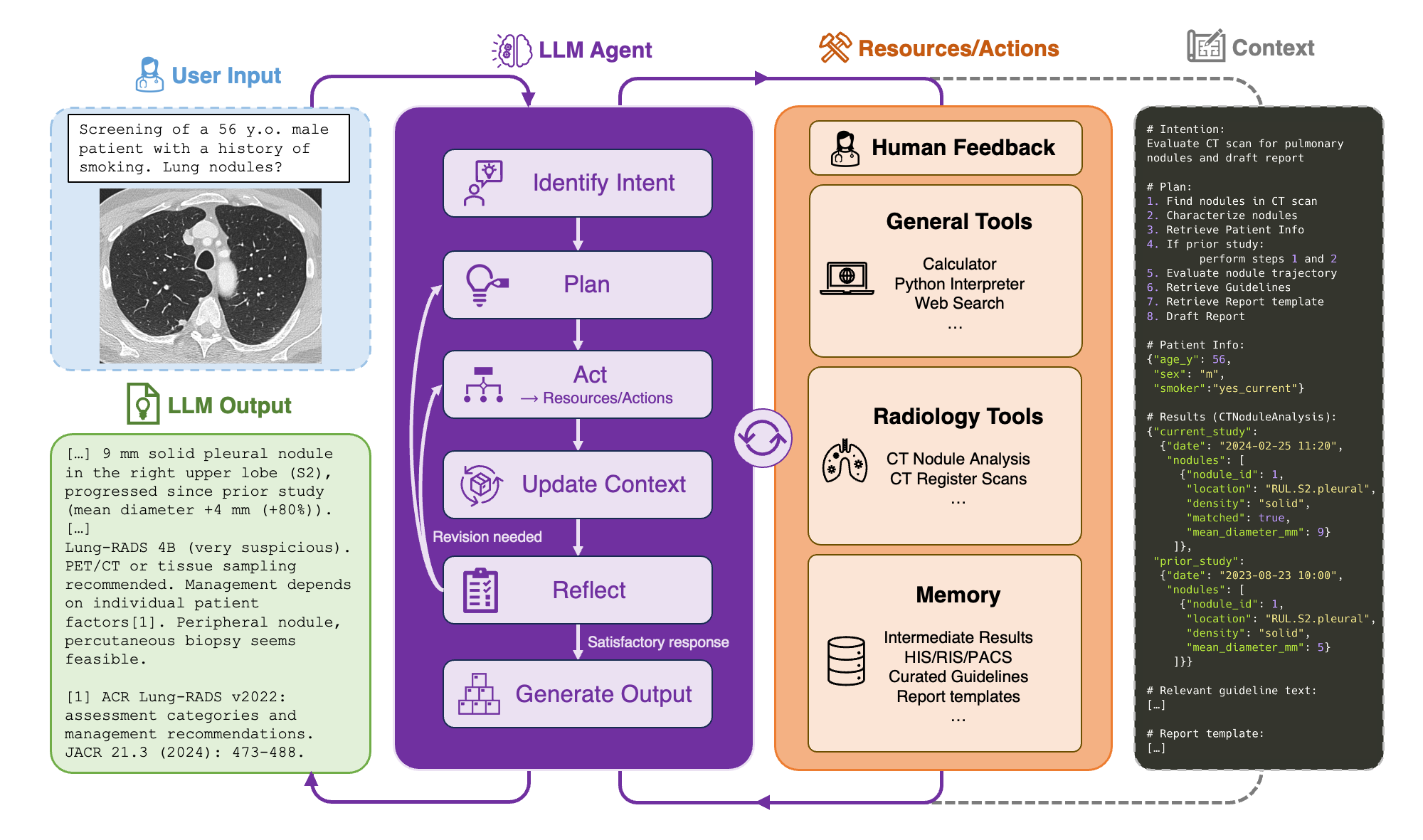}
    \caption{Large language models (LLMs) in agentic workflows. In this example, the user asks for lung nodules in a CT scan (screening setting). An agentic workflow represents an iterative process in which the large language model (LLM) identifies what to do (Identify Intent), conceives a plan how to proceed (Plan), and selects from a set of available actions and resources (Act) including general-purpose tools like calculators, radiology-specific tools (e.g. CT lung nodule analysis software), and other resources like healthcare information systems or databases (e.g., guidelines like Lung-RADS). The LLM iteratively refines its context based on new information (Update Context), revising its plan or performing additional actions if necessary (Reflect). In this example, the identification of a prior study triggers a more refined approach involving analyzing and correlating previous pulmonary nodules with the current study. This process culminates in autonomously generating a comprehensive output tailored to the initial query, incorporating relevant medical guidelines and recommendations. HIS: Hospital Information System. RIS: Radiology Information System. PACS: Picture Archiving and Communication System.}
    \label{fig:3_agentic}
\end{figure}

This will help pave the way for narrow applications, as well as for more complex developments like LLM-based agents capable of autonomously interacting and solving problems in multi-step approaches(Fig.~\ref{fig:3_agentic}), and multimodal models harnessing several input modalities (e.g., images, videos and text) to further add value to the radiology workflow\cite{chenCheXagentFoundationModel2024b,reed2022a}.

While this review focuses on the technical aspects of LLM adaptation, practical applications need to ensure applicable regulatory compliance, such as HIPAA and GDPR, as well as future AI-specific issues like the White House executive order on AI or the EU AI Act. Beware that LLM applications in radiology may be classified as medical devices depending on their impact on clinical decisions (e.g., for triage or diagnostic support), with regulatory requirements varying by region and uses\cite{gilbertLargeLanguageModel2023}.

\newpage
\section{Best Practices for LLMs in Radiology}
LLM research is moving extremely fast, and important developments often outpace scientific peer-review timescales. Drawing from current literature, practical experience, and interdisciplinary discussions, we identify best practices for using and adapting LLMs in radiology. Current reviews operating at varying levels of domain-specificity and technical depth contain further details\cite{bhayanaChatbotsLargeLanguage2024,akincidantonoliLargeLanguageModels2023,omiyeLargeLanguageModels2024,heSurveyLargeLanguage2023,changSurveyEvaluationLarge2023,zhaoSurveyLargeLanguage2023}.

\subsection{Awareness of Challenges and Limitations}

\textbf{Confabulations}. Arguably the biggest hurdle in using LLMs in radiology settings is the propensity to generate plausible-sounding, yet incorrect information (``hallucination'', although ``confabulation'' may be more appropriate\cite{smithHallucinationConfabulationNeuroanatomy2023,hatemChatbotConfabulationsAre2023}). Fabricating information is not limited to text-only LLMs: Multimodal LLMs can hallucinate objects in provided images\cite{liEvaluatingObjectHallucination2023}. Confabulations are aggravated by the LLMs' ineptitude to reliably convey a degree of confidence in their assessments, which is pivotal for relaying relevant medical information\cite{zhaoExplainabilityLargeLanguage2024}. Several techniques like chain-of-thought prompting (instructing the model to break up a problem into intermediary steps, e.g. ``Let's think step-by-step''), in-context learning (i.e., providing examples in the input), self-consistency (repeatedly use the same input to evaluate the consistency of the responses) and RAG can mitigate confabulations\cite{tonmoyComprehensiveSurveyHallucination2024}. Still, due diligence is warranted to reduce the risk of negative outcomes associated with incorrect generations.

\begin{tcolorbox}[colback=gray!10,colframe=gray!20,boxrule=0.5pt,arc=4mm]
\textbf{Recommendation:} Critically evaluate LLM outputs for wrong and potentially harmful statements. Use mitigation techniques like chain-of-thought prompting and retrieval-augmented generation to increase the factual correctness.
\end{tcolorbox}
\vspace{\baselineskip}

\textbf{LLMs are Language Processors, Not Knowledge Databases}. Numerous studies natively queried LLMs and evaluated the answers for their factual correctness, which is a suboptimal use of their abilities. Transformer-based LMs can store knowledge to some extent\cite{petroni2019language}, however, LLMs build on diverse and mostly unverified training sources and may not store reliable healthcare knowledge\cite{truhnLargeLanguageModels2023a,thirunavukarasuLargeLanguageModels2023}. Additionally, without modification (e.g., allowing the LLM to access search engines) or retraining, LLMs are restricted to the data available at the time of their training.

\begin{tcolorbox}[colback=gray!10,colframe=gray!20,boxrule=0.5pt,arc=4mm]
\textbf{Recommendation:} Use LLMs to interpret and synthesize information, while relying on validated knowledge sources for factual, verifiable content. For knowledge-based tasks, carefully evaluate LLM outputs and consider using customized or fine-tuned models validated for specific domains.
\end{tcolorbox}
\vspace{\baselineskip}

\textbf{Open vs. Closed Models}. Several levels of accessibility exist for LLMs. Open LLMs (e.g., BLOOM(27)) are distributed by sharing model weights(i.e., \textit{open-access}) — and associated code (i.e., \textit{open-source}). Closed models provide no access to weights or code. Model access can also be regulated through gating mechanisms, such as hosting closed models (e.g., OpenAI's ChatGPT), or by imposing usage restrictions (e.g., Meta's LLaMA 2 and 3\cite{grattafiori2024llama3herdmodels}). Fully closed LLMs are only accessible to those directly involved with the model (e.g., Med-Gemini\cite{saab2024capabilitiesgeminimodelsmedicine}).

Different benefits and risks exist along this gradient. Closed models provide greater risk control, allowing the developers to monitor inputs and employ usage guardrails. Although both open and closed models can be customizable and HIPAA-/GDPR-compliant, closed models provide lower auditability and transparency, which can also affect downstream reproducibility. They also usually rely on the developer for reliability and data privacy assurances, and may have use case restrictions (such as explicitly excluding the use for processing medical information). Hosted, closed LLMs can be updated without notice, potentially affecting performance and outputs. Awareness of these changes is vital for consistent application.

Open models, on the other hand, enable broader perspectives to use and evaluate the model fully and transparently, and can be hosted and updated on the user's computational infrastructure of choice. However, these open models may often not match performance of closed, hosted models. 

\begin{tcolorbox}[colback=gray!10,colframe=gray!20,boxrule=0.5pt,arc=4mm]
\textbf{Recommendation:} The use of open, locally running models enhances control and is encouraged to increase privacy preservation, reproducibility and customizability for radiological applications. Use closed, hosted models such as ChatGPT with caution, do not input data that you would not be comfortable sharing publicly.
\end{tcolorbox}
\vspace{\baselineskip}

\textbf{Reproducibility and Explainability}. Reproducibility is challenging to achieve with LLMs due to prompt brittleness (the sensitivity to even small changes in the input), the inherent stochasticity of LLMs (by design, the LLM predicts a probability distribution to be sampled from when outputting the next token) and the randomness that becomes apparent at a large scale (when even very rare events such as hardware errors can have an effect). Silent updates to closed source LLMs due to retraining can exacerbate this issue. Within healthcare, interpretable and explainable models are standard within a degree of reason\cite{bradyDevelopingPurchasingImplementing2024}. While current LLMs lack comprehensive interpretability mechanisms, conversational LLMs can provide some transparency by explaining their outputs through dialogue. Additionally, techniques like chain-of-thought prompting can improve model reasoning by encouraging step-by-step problem solving while at the same time allowing the user to see possible intermediate considerations\cite{savageDiagnosticReasoningPrompts2024a}.

\begin{tcolorbox}[colback=gray!10,colframe=gray!20,boxrule=0.5pt,arc=4mm]
\textbf{Recommendation:} Be aware that reproducibility can be limited in LLMs. If reproducibility is a priority, model version, employed hyperparameters and prompt should be documented. Chain-of-thought prompting may improve outputs while providing some limited explainability for LLM outputs.
\end{tcolorbox}
\vspace{\baselineskip}

\textbf{Risk of bias}. LLMs pick up patterns from the training data at all stages of their training, including various types of bias (e.g., associated with gender, race, religion) inherent to the training data\cite{omiyeLargeLanguageModels2024,gallegosBiasFairnessLarge2023}. For instance, an LLM instructed to complete the sentence ``the doctor is a 35 year old'' may predict ``man'' significantly more frequently than ``woman'' whereas ``the nurse is a 35 year old'' can lead to opposite results\cite{thakurUnveilingGenderBias2023}. Other forms of bias affecting LLM integration include clinical confounding bias, technical bias and automation bias\cite{bradyDevelopingPurchasingImplementing2024}. These biases can be partially mitigated through careful curation of training and fine-tuning data and through alignment efforts\cite{koccakbias}.

\begin{tcolorbox}[colback=gray!10,colframe=gray!20,boxrule=0.5pt,arc=4mm]
\textbf{Recommendation:} Be aware that current LLMs can contain and propagate several kinds of biases. When creating a custom LLM, careful curation of pretraining and fine-tuning data, alignment and appropriate evaluation strategies can help mitigate biases.
\end{tcolorbox}
\vspace{\baselineskip}

\textbf{Data privacy and security concerns}. Data privacy challenges stem from both the training and inference phases of LLMs. Training LLMs can put sensitive health information at risk of being regurgitated if not adequately anonymized, as such data can potentially be extracted from LLMs after training\cite{carliniExtractingTrainingData2021}. Additionally, during a chat session, the prompt context is not ``forgotten'', and potentially sensitive input data can reappear within the same session\cite{priyanshuAreChatbotsReady2023}. It is recommended to de-identify data before using LLMs, although this standard can be challenging to achieve.

The versatility of LLMs has another downside: even aligned LLMs are susceptible to prompt-based attacks that elicit unwanted behavior (jailbreaking), such as revealing sensitive information or providing harmful content. New exploits are discovered frequently and large organizations like OpenAI employ red-teaming (i.e., deliberately provoking or stress-testing the LLM with challenging inputs to discover its limitations and potential risks or unwanted behavior) and offer ``bug bounties''\cite{AnnouncingOpenAIBug}. LLMs that can handle image inputs often include an optical character recognition (OCR) functionality, which allows them to read and interpret text in the image, rendering manipulated images a potential Trojan horse for adversarial attacks\cite{bagdasaryanAbusingImagesSounds2023}.

\begin{tcolorbox}[colback=gray!10,colframe=gray!20,boxrule=0.5pt,arc=4mm]
\textbf{Recommendation:} Be aware that LLMs can store and regurgitate training data verbatim. To ensure privacy, exclude private data from training datasets and inputs, especially when using hosted models, and prefer locally running models if possible. Be aware that the flexibility of LLMs enables innovative adversarial methods.
\end{tcolorbox}
\vspace{\baselineskip}

\textbf{Creating and Adapting LLMs}. Training LLMs on domain-specific data, like radiology reports and images, can enhance the performance on downstream tasks\cite{chenCheXagentFoundationModel2024b,vanveenClinicalTextSummarization2023}. Detailed technical best practices for LLM adaptation are case-specific, rapidly evolving and are discussed in more detail in the literature\cite{heSurveyLargeLanguage2023,zhaoSurveyLargeLanguage2023}.

The high level of computer and machine learning engineering expertise required for LLM training makes close collaboration with computer scientists essential for developing, monitoring and deploying LLMs. Radiologists, on the other hand, should not be relegated to the role of data providers. Instead, they should contribute their domain expertise by leading the development of these powerful tools, defining specific objectives, contributing to data set design and conducting thoughtful and domain-oriented evaluation.

\begin{tcolorbox}[colback=gray!10,colframe=gray!20,boxrule=0.5pt,arc=4mm]
\textbf{Recommendation:} Domain-adapting LLMs can increase the performance on specific tasks and can be a viable strategy when aiming for high levels of customization. Creating LLMs for radiology is a team effort between radiologists and computer scientists.
\end{tcolorbox}
\vspace{\baselineskip}

\textbf{Pretraining LLMs}. Pretraining modern large LLMs exceeding a billion parameters demands immense resources, including massive datasets\textsuperscript{1}, substantial infrastructure\textsuperscript{2}, and technical expertise. High costs leave little room for trial and error or even repeated iteration at scale and necessitate meticulous planning. Key decisions involve choosing the right model architecture, size, compute cluster setup\cite{anthonyCaseCoDesigningModel2024,rajbhandariZeROMemoryOptimizations2020,soldainiDolmaOpenCorpus2024}, and meticulously curating datasets with trillions of tokens while ensuring data quality through content filtering, deduplication, and optimal tokenization methods\cite{zhaoSurveyLargeLanguage2023,soldainiDolmaOpenCorpus2024}. Detailed best pretraining practices constitute a volatile area of research, and are discussed in recent literature\cite{zhaoSurveyLargeLanguage2023}.

Fine-tuning pretrained LLMs is much less resource-intensive\textsuperscript{3} and should be prioritized. The model size should balance the necessary capacity to fulfill envisioned tasks and computational resources and ecological footprint. The LLaMA 2 model family is a prominent example of open-source pretrained LLMs available in different sizes (7B, 13B, 70B parameters) that has spawned successful ``laminoid'' models like Vicuna\cite{zhengJudgingLLMasaJudgeMTBench2023}, Alpaca\cite{alpaca} and the multimodally capable LLaVa\cite{liu2023visual}. With a few notable exceptions\cite{singhalExpertLevelMedicalQuestion2023a,jiangMistral7B2023}, larger LLMs tend to perform better across evaluations than their smaller-scale counterparts, at the cost of higher memory and compute requirements. Smaller \textit{fine-tuned} models may outperform larger general-purpose LLMs\cite{fuTinyTitansCan2024}. Recently, alternatives to the predominant single decoder transformer models have become increasingly popular, such as mixture-of-expert (MoE, e.g. Mistral)\cite{jiangMixtralExperts2024}, and structured state space models\cite{guMambaLinearTimeSequence2023}.

\begin{figure}
    \centering
    \includegraphics[width=1.0\linewidth]{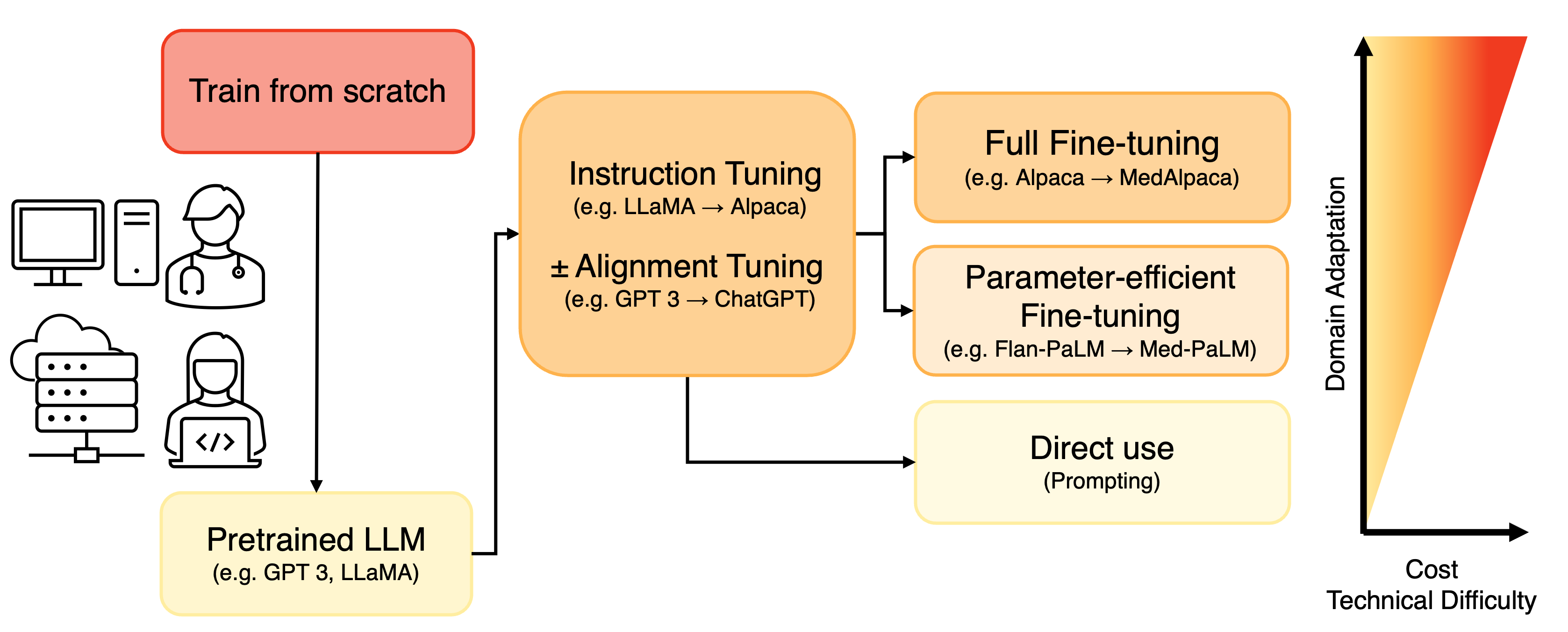}
    \caption{Simplified overview of training and domain adaptation methods for large language models (LLMs). Initially, any LLM has to incorporate general language understanding through self-supervised pretraining on vast amounts of raw text data. This step is technically challenging and resource-intensive. Starting with an already pretrained, instruction-tuned or even aligned LLM is easier. These aligned LLMs can be used directly via prompting. Should a task require specialization beyond what can be added via context (e.g. to introduce refined radiological terminology and tasks), fine-tuning through custom instruction tuning datasets can help. Full fine-tuning multi-billion parameter models is resource-intensive; in resource-constrained environments, parameter-efficient techniques like low-rank adaptation (LoRA, a method that updates only a small number of parameters instead of the full model to reduce computational costs) can facilitate fine-tuning endeavors. All domain-adaptation strategies benefit from a close collaboration of domain experts (radiologists), computer scientists and machine learning engineers.}
    \label{fig:4_domain_adaptation}
\end{figure}

\begin{tcolorbox}[colback=gray!10,colframe=gray!20,boxrule=0.5pt,arc=4mm]
\textbf{Recommendation:} Projects aiming to create radiology LLMs should avoid costly, technically challenging pretraining and attempt using and adapting open, general-purpose LLMs. The choice of model and approach should strike a balance between task-specific needs and available resources.
\end{tcolorbox}
\vspace{\baselineskip}

\footnotetext{$^1$ GPT 3 was trained on 0.5 trillion tokens\cite{shahCreationAdoptionLarge2023}, LLaMA 2 was trained on 1.4 trillion tokens\cite{touvronLlamaOpenFoundation2023}. 1 trillion tokens is roughly the equivalent of 15 million books\cite{huyenRLHFReinforcementLearning2023a}. A typical chest x-ray report is in the range of 200 tokens (GPT 2 tokenizer, MIMIC-CXR reports).}

\footnotetext{$^2$ Pretraining a smaller BERT model (109 million parameters) on clinical documents required 3 weeks of continuous training on 24 NVIDIA A100 GPUs, equating to 24 x (3 x 7 x 24h) x \$33/8 GPUs/h (typical current A100 on-demand price) = \$ 49,896 for a single pretrained model\cite{savageDiagnosticReasoningPrompts2024a}. GPT3 (175 billion parameters) is estimated to have been trained for \$4.6 millions\cite{shahCreationAdoptionLarge2023}.}

\footnotetext{$^3$ DoctorGLM is an LLM fine-tuned (with LoRA) on 100k QA pairs, trained for 3.75 hours at \$5/h, amounting to approximately 18.75 USD\cite{xiongDoctorGLMFinetuningYour2023}. LLaVA-Med: 8 A100 GPUs for 15h (495 USD at \$33/8 GPUs/h)\cite{liLLaVAMedTrainingLarge2023}.}

\textbf{Instruction tuning}. Starting from a pretrained LLM, models can be fine-tuned in a supervised way to increase the ability to follow instructions (instruction tuning) and improve their performance on downstream tasks\cite{howard2018FineTunedLanguage}. An instruction-following dataset can be constructed with instances following a pattern like \{instruction(+input)\}\{output\}\cite{alpaca}. These instances can be created from paired sources (e.g. medical exams questions, flashcards\cite{hanMedAlpacaOpenSourceCollection2023}), by restructuring existing datasets (e.g. CheXinstruct\cite{chenCheXagentFoundationModel2024b}) or synthesized by another LLM\cite{alpaca}. LLMs perform best on tasks appearing frequently in their training data. Balancing instruction datasets is therefore relevant to radiology and other medical specialties who deal with the long tail problem\cite{kahnLongTail} and can mitigate overfitting on certain tasks. Non-English LLM users benefit from the insight that English-language instruction tuning can also increase an LLM's task-specific capabilities in other languages\cite{muennighoffCrosslingualGeneralizationMultitask2023}. The typical size of an instruction tuning dataset is in the range of thousands of examples (e.g. 52k in Alpaca), while already tens of examples can improve performance on certain tasks\cite{singhalExpertLevelMedicalQuestion2023a}. For instruction tuning, ensuring high data quality is much more important than during pretraining. Data quality and diversity increase the model's generalization ability and take priority over dataset size\cite{zhaoSurveyLargeLanguage2023}. Instruction tuning can boost domain- and task-specific performance of LLMs across several dimensions, e.g. accuracy, robustness, and fairness\cite{liangHolisticEvaluationLanguage2023}. Fine-tuned, instruction-following models can be used directly or further fine-tuned with a similarly structured dataset.

\begin{tcolorbox}[colback=gray!10,colframe=gray!20,boxrule=0.5pt,arc=4mm]
\textbf{Recommendation:} Instruction tuning can be used for domain- and task-adaptation. The instruction dataset should be carefully curated, balancing domain- and task-specific needs, prioritizing quality and diversity over dataset size.
\end{tcolorbox}
\vspace{\baselineskip}

\textbf{Aligning LLM Outputs with Human Preferences}. Following instruction tuning, alignment tuning seeks to use human feedback on generated outputs to align an LLM with human preferences. In RLHF, an instruction-tuned model generates multiple naïve outputs for the same prompt, which are then ranked by humans according to their preferences. This ranking data is used to train a reward model, which is in turn used to inform further fine-tuning of the LLM towards following human preferences. Curating large high-quality preference datasets from human annotators is costly, and successfully training an reinforcement learning (RL) model is considerably more challenging than instruction tuning\cite{rafailovDirectPreferenceOptimization2023}. Newer approaches aim to obviate the RL step in lieu of simpler objectives and facilitate collecting alignment data by requiring less complex decisions (``which output is preferred?'' or ``is this output good?'')\cite{rafailovDirectPreferenceOptimization2023,ethayarajhKTOModelAlignment2024}. Other promising research uses LLMs for generating and evaluating the quality of supervised training datasets\cite{liSelfAlignmentInstructionBacktranslation2023,wang-etal-2023-self-instruct}, or for validating the accuracy of LLM outputs\cite{geroSelfVerificationImprovesFewShot2023}.

\begin{tcolorbox}[colback=gray!10,colframe=gray!20,boxrule=0.5pt,arc=4mm]
\textbf{Recommendation:} If further domain- and task-specific adaptation is necessary, consider alignment tuning with human feedback data. This adjusts the LLM to match human preferences, such as adhering to departmental reporting styles.
\end{tcolorbox}
\vspace{\baselineskip}

\textbf{LLMs on a budget}. Parameter-efficient techniques that approximate the performance of full fine-tuning while only updating a much smaller number of parameters can facilitate LLM adaptation. LoRA (low-rank adaptation) is a popular method which introduces trainable rank decomposition matrices into the LLM, while the original model is kept frozen\cite{huLoRALowRankAdaptation2021}. It is possible to fine-tune several low-rank matrices for specific tasks, while keeping just one copy of the large main LLM, thereby greatly reducing memory requirements. Another approach trains task-specific prompt vectors that are introduced alongside the rest of the prompt (soft prompt tuning)\cite{singhalLargeLanguageModels2023,lesterPowerScaleParameterEfficient2021}. Proxy tuning enhances the capabilities of a larger-scale LLM by integrating the differences in token predictions between two smaller models with the predictions of the larger LLM\cite{liuTuningLanguageModels2024}, rivaling the more expensive process of fine-tuning the larger LLM directly in performance. PEFT is a popular library implementing several parameter-efficient techniques\cite{mangrulkar2022PEFT}.

Quantization is the process of mapping higher-bit values to lower-bit approximations (e.g. converting models weights or activations from 16-bit float values to 8-bit integers). Quantization entails a quantization error due to rounding operations, but reduces memory requirements and increases inference speed. QLoRA is a method that combines the benefits of LoRA and quantization\cite{dettmersQLoRAEfficientFinetuning2023}. Distillation techniques can significantly decrease the requirements for hosting and running LLMs and thus facilitate deployment in constrained environments. These methods aim at reducing training data\cite{wangDatasetDistillation2020} or model size, for instance by training smaller models (``student'') to replicate the behavior of larger LLMs (``teacher'')\cite{hsiehDistillingStepbyStepOutperforming2023}. All presented methods have a trade-off in performance\cite{dettmersQLoRAEfficientFinetuning2023}, but make it possible to run smaller LLMs with reasonable performance on desktop computers\cite{Ollama}.

\begin{tcolorbox}[colback=gray!10,colframe=gray!20,boxrule=0.5pt,arc=4mm]
\textbf{Recommendation:} Parameter-efficient fine-tuning methods like LoRA and quantization can significantly reduce the computational costs of domain-adaptation and inference.
\end{tcolorbox}
\vspace{\baselineskip}

\textbf{Multimodality}. While a variety of approaches to implement multimodality exists\cite{huangMultimodalFoundationModels2024}, here, we refer to models with an LLM backbone that can process inputs from $\geq$2 data sources (e.g., images and text), mirroring the setting that is common in radiology, as \textit{large multimodal models}(LMMs). It is generally expected that fine-tuned LMMs will play an increasing role in radiology over the next few years, but so far, general-domain LMMs dominate the literature\cite{zhangMMLLMsRecentAdvances2024} and can sometimes outperform domain-specific models\cite{noriCanGeneralistFoundation2023}. Most notably, GPT-4V is a closed LMM that has shown promising yet flawed performance on analyzing radiological images\cite{yangDawnLMMsPreliminary2023,91}. Med-PaLM M is a closed healthcare-centered LMM evaluated for radiology report generation, but no further radiology-specific benchmarks are documented\cite{tuGeneralistBiomedicalAI2023}. CheXagent is a recently introduced imaging modality-specific open LMM outperformed several other general and medical domain LLMs on a set of CXR-specific tasks\cite{chenCheXagentFoundationModel2024b}. Other early examples of radiology LMMs include RadFM\cite{wuGeneralistFoundationModel2023}, RaDialog\cite{pellegriniRaDialogLargeVisionLanguage2023}, CXR-LLaVA\cite{leeCXRLLAVAMultimodalLarge2024}. LMMs typically require images as part of their instruction tuning set\cite{zhangMMLLMsRecentAdvances2024}.

\begin{tcolorbox}[colback=gray!10,colframe=gray!20,boxrule=0.5pt,arc=4mm]
\textbf{Recommendation:} Keep an eye on large multimodal models, which - unlike text-only LLMs - incorporate multiple types of data modalities (e.g., images and text). This extends their potential applications in radiology but comes with added complexity.
\end{tcolorbox}
\vspace{\baselineskip}

\subsubsection{Model Cards and Versioning}
\textbf{Model Cards and Versioning.} Regardless of whether the model is shared or solely used within the department, model cards detailing training datasets, evaluation metrics, and known limitations such as biases should be used to promote transparency, informed usage and reproducibility\cite{touvronLlamaOpenFoundation2023,96}. Intermediate model checkpoints should be saved, assigned a version number and documented as they can give insights about the models scaling behavior\cite{bidermanPythiaSuiteAnalyzing2023} and can be used to return to a previous model state (e.g., when observing performance drops post training).

\begin{tcolorbox}[colback=gray!10,colframe=gray!20,boxrule=0.5pt,arc=4mm]
\textbf{Recommendation:} Use model cards and versioning when creating, adapting and sharing LLMs.
\end{tcolorbox}
\vspace{\baselineskip}

\subsection{Evaluating LLMs}
Despite general language understanding, generation and reasoning skills, LLMs may lack proficiency in specialized domains like radiology, which have unique terminology, concepts, and tasks. Single metrics can fall short in evaluating complex tasks\cite{goodmanAIGeneratedClinicalSummaries2024}. Thus, radiology LLM evaluation frameworks should cover domain-specific tasks and include expert human assessment.

Automatic scoring metrics for conditional text generation assess LLM responses in comparison to (human) references. These metrics measure the overlap of sentence parts (BLEU\cite{papineni2002bleu}, ROUGE-L\cite{lin2004rouge}), the similarity of medical concepts (e.g., MEDCON\cite{yimACIBENCHNovelAmbient2023}), or transformer-based similarity with general (BERTScore\cite{zhang2019bertscore}) and domain-specific focus (F1CheXBERT\cite{zhangOptimizingFactualCorrectness2020}, RadGraph Score\cite{delbrouckImprovingFactualCorrectness2022a}). Another method is to transform complex tasks like question-answering (QA) into multiple-choice questions (e.g. PubMedQA\cite{jin2019pubmedqa}), and measure the LLM's accuracy in selecting the correct answer.

Single metrics often fall short in evaluating complex, open-ended tasks like QA and summarization\cite{goodmanAIGeneratedClinicalSummaries2024}, necessitating benchmark frameworks for comprehensive model assessment\cite{liangHolisticEvaluationLanguage2023,hendryckstest2021,wang2018glue}. Radiology texts, with their specific terminology and general anchoring on medical imaging data, require tailored benchmarks. Early examples like RaLEs\cite{chavesRaLEsBenchmarkRadiology2023} evaluate LLMs on language radiology-specific tasks, such as anatomical relationship extraction, procedure selection and report summarization, while RadBench\cite{wuGeneralistFoundationModel2023} focuses on multimodal tasks including visual QA and report generation. Despite the need for public evaluation frameworks for transparency and reproducibility, concerns arise that such data might inadvertently be used for training, skewing results\cite{zakkaAlmanacRetrievalAugmentedLanguage2024}. Gated access to evaluation data may be an option to prevent data contamination.

Human expert evaluation is paramount for assessing the real-world correctness, conciseness, and completeness of outputs, but hard to scale, and may need calibration for consistent feedback. Notably, medical expertise exists on a spectrum; for instance, clinicians, medical students, radiology residents and board-certified or subspecialized radiologists may all focus on different aspects of a report, affecting their interpretation. This warrants a careful selection of readers and documentation of experience level.

Effective evaluation for LLMs in radiology could be constructed as a multi-step framework harnessing the qualities of all presented methods: automatable metrics (single or as part of a benchmark) are used to compare, optimize and select different approaches to get an optimal configuration, which is subsequently evaluated by human experts\cite{vanveenClinicalTextSummarization2023}. Aside from task-specific performance, outputs should be evaluated for criteria like potential harmfulness. Evaluation should continue after deployment, as updates to the LLM or shifts in input data could lead to unexpected and potentially harmful outputs.

\begin{tcolorbox}[colback=gray!10,colframe=gray!20,boxrule=0.5pt,arc=4mm]
\textbf{Recommendation:} For safe and effective use, LLMs require thorough evaluation during and after development. Evaluation should include general capabilities as well as radiology-specific benchmarks and include evaluation by radiologists. Evaluation datasets should be handled responsibly, as data contamination can lead to overly optimistic evaluations.
\end{tcolorbox}

\subsection{Using and Augmenting LLMs}

\textbf{When not to use an LLM}. LLMs have substantially democratized NLP research, leading to a steep increase in publications and peer-review work. The versatility of LLMs makes it attractive to conduct more and more refined feasibility studies. However, building \textit{and} using LLMs are resource-intensive and carry considerable ecological footprints\cite{truhnEcologicalFootprintMedical2023,pattersonCarbonEmissionsLarge}. While transfer learning hopes to mitigate train time, continuous training will also have an effect on CO\textsubscript{2} emissions\cite{dooEvaluationClimateAwareMetrics2023}. Using an LLM for tasks that could be solved more efficiently with simpler approaches (for instance, by using a smaller LM or through employing regular expressions for traditional NLP tasks) constitutes over-engineering and should be avoided.

\begin{tcolorbox}[colback=gray!10,colframe=gray!20,boxrule=0.5pt,arc=4mm]
\textbf{Recommendation:} Critically evaluate if using an LLM can be substituted by a smaller language model or a more efficient method. Avoid (research) waste.
\end{tcolorbox}
\vspace{\baselineskip}

\textbf{Prompting LLMs}. A distinctive and revolutionary feature of modern LLMs is the direct applicability to various downstream tasks without the need to retrain the model. Most radiologist users will interact with an LLM through prompting. Accurately instructing the model is often enough to get good results with current general-domain LLMs. However, creating and refining an effective prompt can be challenging\cite{zamfirescu-pereiraWhyJohnnyCant2023}. Prompt engineering is the process of designing and iteratively adjusting instructions to optimize the LLM output for the envisioned task, as even little variations can lead to significantly different outputs (prompt brittleness). Some good practices for designing effective prompts have emerged\cite{weng2023prompt,OpenAIPlatform,PromptEngineeringGuide}(Table~\ref{tab:prompting_techniques}). The adherence to a task can be increased by providing examples (few-shot prompting, a form of in-context learning with usually 2-5 examples), which can be seen as implicit fine-tuning without updating the model weights\cite{brownLanguageModelsAre2020}. While many LLMs can be prompted to use a specific output format directly (e.g. Python code or JSON\cite{adamsLeveragingGPT4Post2023}), providing a template or output schema (e.g., using Pydantic\cite{samuel_colvin_2024_10619320}) can greatly increase the adherence to such a request.

Apart from the prompt itself, outputs can be influenced by how the next token is sampled. Modern LLMs often use nucleus sampling, which is parameterized by temperature and top-p\cite{holtzmanCuriousCaseNeural2020}. Temperature rescales the probability mass function, affecting the probability of sampling lower-probability tokens, while top-p truncates the distribution to only include the smallest set of tokens whose cumulative probability exceeds p. Lower values for temperature and top-p push the output towards the most probable tokens, decreasing diversity. Setting the temperature to 0 makes the model theoretically deterministic (i.e., the most probable token is always picked); however, in practice, random effects can still influence the LLM output, so perfect reproducibility should not be taken for granted. For most tasks in radiology, when conciseness and the use of specific expressions are a priority over diverse outputs, lower temperature values will likely be preferred, as demonstrated in a clinical text summarization study\cite{vanveenClinicalTextSummarization2023}. Frequency and presence penalty are parameters preventing the LLM from repeating itself too often. For radiology report generation, where certain expressions (e.g., ``lesion'', ``nodule'') are often suitable more than once, and diversity of expression is not a priority, lower penalty values may be better.

\begin{tcolorbox}[colback=gray!10,colframe=gray!20,boxrule=0.5pt,arc=4mm]
\textbf{Recommendation:} To enhance LLM outputs, employ task-specific iterative prompt refinement, provide examples, and adjust hyperparameters like temperature. Document model version, prompts, and parameters used, as outcomes can vary even across different versions of the same LLM.
\end{tcolorbox}
\vspace{\baselineskip}

\begin{table}[htbp]
\renewcommand{\arraystretch}{1.3}
\caption{Prompting techniques and best prompting practices to guide a large language model (LLM)}
\label{tab:prompting_techniques}
% \vspace{1mm}
\begin{tabular}{|p{0.078\textwidth}|p{0.4\textwidth}|p{0.455\textwidth}|}
\hline
\textbf{Category} & \textbf{Recommendation} & \textbf{Example} \\
\hline
\multirow[t]{2}{=}{General} & Using English instructions, even in multilingual models, can increase the output quality. & \\
\cline{2-3}
& Avoid negations, as they may elicit opposite behavior. & ``Answer briefly and concisely'' is better than ``Do not write a verbose text'' \\
\hline
\multirow[t]{2}{=}{Prompt Structure} & Use a suitable input format - this depends on the LLM. Special tokens or delineating strings may be necessary, depending on the LLM. & \texttt{<s>[INST] \textit{Instruction text} [/INST]</s>} 
(example: Mistral-7B-v0.1) \\
\cline{2-3}
& Provide structured inputs (e.g., JSON (JavaScript Object Notation), or through paragraphs with headlines and separating characters). This can increase adherence to the prompt. & \texttt{\{"report": "Findings: Lungs are well expanded. No pleural[...]",\newline
"impression": ``No acute cardiopulmonary process."\}} \\
\hline
\multirow[t]{8}{=}{Prompt Content} & Instruct the model to act as a domain expert. This can increase domain-specific performance. & \texttt{"You are a radiologist subspecialized in thoracic imaging."} \\
\cline{2-3}
& Dictate a style to shape the output language. & \texttt{"Use a scientific tone, concise phrasing and radiological terminology."} \\
\cline{2-3}
& Dictate an output format. Defining output schemas and including a validation step can greatly increase adherence to the request. & \texttt{"Output a list of bullet points."} \\
\cline{2-3}
& Chain-of-thought (CoT) prompting: Guide the model through intermediate reasoning steps before reaching a conclusion (e.g., by instructing to "think step-by-step"). This can boost performance for complex tasks. & \texttt{"A patient presents with shortness of breath and has a history of deep vein thrombosis. What would be an appropriate next step? Let's think step-by-step."} \\
\cline{2-3}
& Self-consistency: Prompt the model several times with the same instructions and identify the most consistent answer. This can boost the performance on reasoning tasks. & Answer 1: ``[...]The answer is 5''\newline
Answer 2: ``[...]The answer is 3''\newline
Answer 3: ``[...]The answer is 5'' 
→ 5 is most likely. \\
\cline{2-3}
& Instruct as detailed as possible. If applicable, explicitly mention necessary intermediate tasks. & ``Help me create a histogram for a scientific publication, illustrating the distribution of document lengths, with the x axis indicating the number of tokens per document and the y axis the number of documents in the respective bin. Use automatic binning. The input is a column vector where each item is the length of one document. Suggest a Python code snippet that can create such a plot.'' \\
\cline{2-3}
& Provide additional context, e.g. through retrieval-augmented generation. This often substantially improves the quality of the answer and reduces the tendency to confabulate. & [Example: Query about incidental pulmonary nodules]\newline
→ Add text of Fleischner Society recommendations for incidental nodules to prompt. \\
\cline{2-3}
& Few-shot prompting: Provide examples from the same task space (inputs) and label space (outputs), with the most similar example at the end. More complex and diverse examples are preferable. This helps the LLM to conform with the task. & \texttt{"Summarize the chest radiograph report by extracting abnormal findings. If there are none, state that there is no acute cardiopulmonary process.\newline\newline
\# Examples:\newline
"report": "Findings: Lungs are well expanded. No pleural[...]",\newline
"impression": ``No acute cardiopulmonary process.''\},"}\newline
[3-5 more examples] \\
\hline
\end{tabular}
\vspace{1mm}
Note that the effectiveness of some recommendations depends on the model, and best practices may change over time as more robust or capable models are being released.
\end{table}

\textbf{Enriching prompts with external information}
Retrieval-augmented generation (RAG) enhances LLM outputs by adding relevant context to user queries\cite{lewisRetrievalAugmentedGenerationKnowledgeIntensive2021}. RAG pairs the LLM with a retrieval module that sources information from databases or curated sites, matching content to queries by similarity. This involves converting chunked documents into vector representations (embeddings) and retrieving and ranking them based on similarity. Domain-specific retrievers such as MedCPT may increase retrieval performance\cite{jinMedCPTContrastivePretrained2023}. Alternatively, additional information can be retrieved from external tools (e.g., calculators). Retrieved information can then be concatenated with the original prompt and used as input for the LLM. An integrated example of this approach in a healthcare setting is presented with Almanac\cite{zakkaAlmanacRetrievalAugmentedLanguage2024}. Using RAG with curated, high-quality sources considerably boosts the factual correctness and quality of LLM outputs over natively prompting the LLM.

\begin{tcolorbox}[colback=gray!10,colframe=gray!20,boxrule=0.5pt,arc=4mm]
\textbf{Recommendation:} Additional information retrieved from sources outside the LLM should be provided as it can greatly increase the factual correctness, the interpretability and the general abilities of LLM-based systems.
\end{tcolorbox}
\vspace{\baselineskip}

\textbf{Balancing prompting, retrieval-augmented generation and fine-tuning.} Choosing the right adaptation method hinges on balancing specific application requirements against constraints like cost, complexity, and the expected level of domain-specific performance\cite{ovadiaFineTuningRetrievalComparing2024}. Full fine-tuning is best suited for applications with deep customization needs but is the most complex and resource-intensive adaptation approach. Parameter-efficient techniques (e.g., LoRA) are more cost-effective for applications requiring less customization and computational resources.

Prompting will only moderately influence concrete radiology-specific abilities but can unlock reasoning behavior (through chain-of-thought prompting) and guide the LLM to the right task space and output format (through the provision of typically 2-5 examples in the sense of few-shot prompting). It is the method of choice for scenarios requiring minimal computational expense and rapid development, when speed is more critical than customization. For many tasks, general-purpose LLMs will deliver a decent and sufficient performance. On the other hand, few-shot prompting requires 1) a number of examples for a given task, 2) fitting these examples within the limited context length of the LLM, and 3) a longer prompt also increases the cost when using closed-source LLMs billed by the token (e.g., ChatGPT).

RAG provides additional context to the LLM and shines in scenarios where dynamically incorporating up-to-date information or external content not present in the model's original training data is beneficial. Apart from increasing the quality of the output, this significantly diminishes the LLM's propensity to confabulate and enhances explainability by enabling users to verify the output (provided that the automatically retrieved information is cited or otherwise made available to them). As an in-context method, RAG has similar restrictions imposed by context window length and token usage.

\begin{figure}
    \centering
    \includegraphics[width=1.0\linewidth]{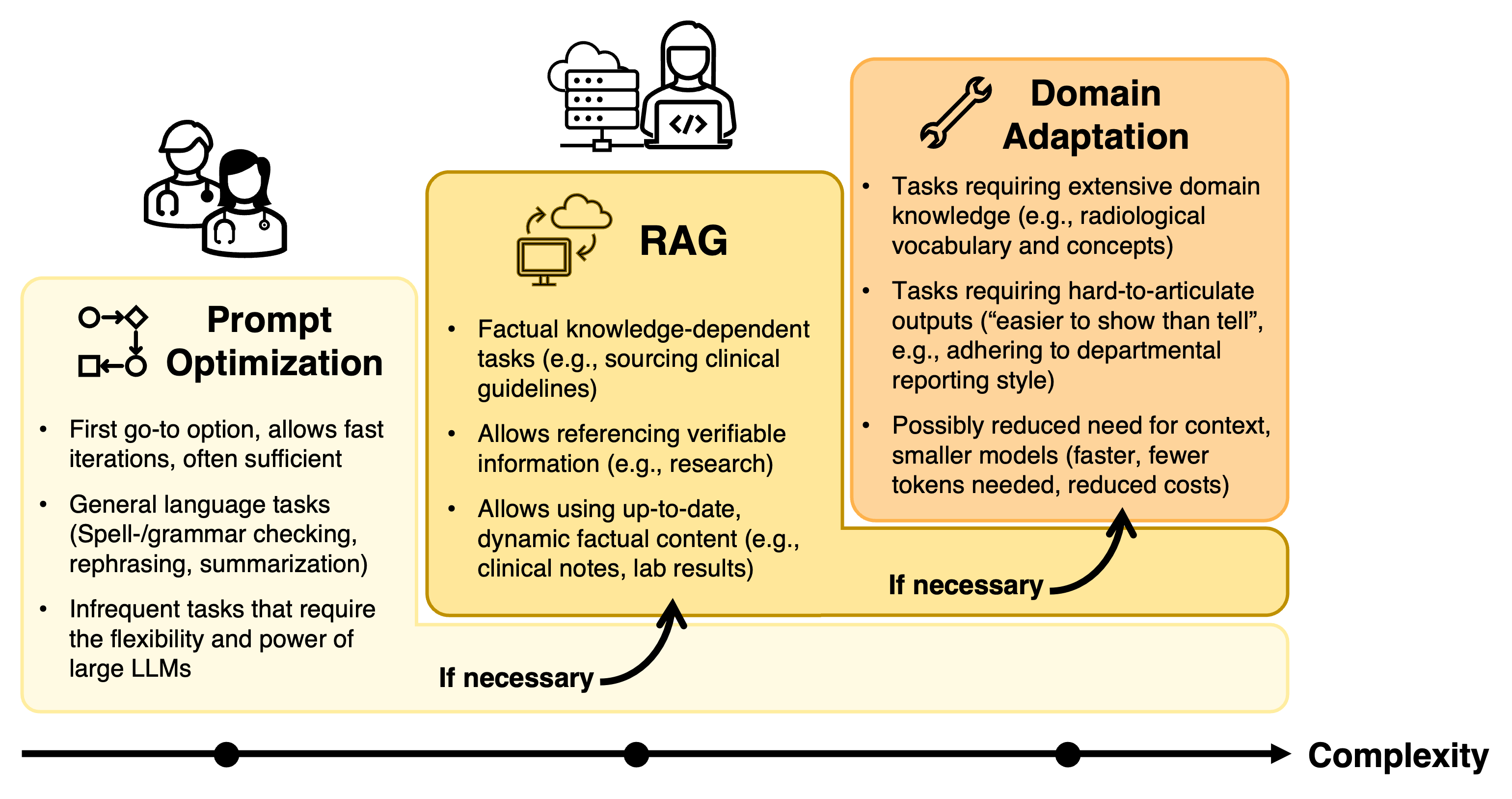}
    \caption{Iterative LLM Optimization for Radiology Tasks. When applying a large language model (LLM) to a new problem, users should start by iteratively refining the input prompt with task-specific instructions. Techniques like role suggestion (e.g., "act as a diagnostic radiologist") or chain-of-thought prompting (asking the model to break down its reasoning process into explicit steps) can often improve results. Especially for knowledge-dependent tasks, Retrieval-Augmented Generation (obtaining additional information from external sources) effectively augments performance (e.g., by incorporating guideline texts or dynamic clinical information). Fine-tuning the model with task-specific data is the most resource-intensive optimization step, as the model is permanently modified towards more specialized radiological applications, requiring additional training data. The presented method groups increase in complexity but are not mutually exclusive; prompt engineering remains instrumental for retrieval-augmented generation-augmented and fine-tuned models.}
    \label{fig:5_balancing_methods}
\end{figure}

\begin{tcolorbox}[colback=gray!10,colframe=gray!20,boxrule=0.5pt,arc=4mm]
\textbf{Recommendation:} Use an iterative approach starting with prompt optimization and retrieval-augmented generation with general-domain LLMs, and explore fine-tuning if needed, to optimize performance for radiology tasks.
\end{tcolorbox}

% ===== TABLE

\begin{table}[htbp]
\renewcommand{\arraystretch}{1.3}  % Increases the spacing between rows
\caption{Best practices for using and adapting large language models (LLMs) in radiology.}
\label{tab:best_practices}
\vspace{1mm}
\begin{tabular}{|p{0.15\textwidth}|p{0.7\textwidth}|>{\centering\arraybackslash}p{0.08\textwidth}|}
\hline
\textbf{Category} & \textbf{Best Practice} & \textbf{Difficulty} \\
\hline
\multirow[t]{4}{*}{General} & Be aware of limitations like confabulations (plausible-sounding, but incorrect statements), biases, data privacy and reproducibility issues, adversarial attacks, ethical implications of using an LLM for medical tasks, and the use of LLMs as medical devices. & \textbullet/\textbullet\textbullet \\
\cline{2-3}
& Utilize LLMs as language processors together with validated knowledge sources, and carefully evaluate their outputs when used for knowledge-based tasks. & \textbullet \\
\cline{2-3}
& Critically evaluate if an LLM is needed for the task at hand. Avoid over-engineering and (research) waste. & \textbullet \\
\cline{2-3}
& For enhanced control, prefer local, open LLMs (i.e., models with publicly available weights, training code, and documented training procedures) over cloud-hosted, closed LLMs (i.e., models where architecture, weights, and training details are not publicly accessible). & \textbullet\textbullet \\
\hline
\multirow[t]{3}{=}{Creating and Adapting LLMs} & Start with prompt optimization (refining input instructions) before considering more resource-intensive approaches. Only proceed to fine-tuning (retraining the model on specific data) when simpler techniques prove insufficient - such as prompt optimization, in-context learning (providing examples in the prompt), or retrieval-augmented generation (enhancing responses with external information). & \textbullet\textbullet \\
\cline{2-3}
& Pretraining LLMs (training from scratch on massive amounts of text) can achieve maximum domain adaptation but should be avoided in most cases due to its costs and technical challenges. & \textbullet\textbullet\textbullet\textbullet \\
\cline{2-3}
& Use curated data for instruction tuning for domain adaptation and improvement on specific tasks. Use alignment tuning for further domain adaptation and increasing adherence to human preferences. & \textbullet\textbullet/\textbullet\textbullet\textbullet \\
\hline
\multirow[t]{2}{=}{Evaluating LLMs} & LLMs need comprehensive, meticulous and ongoing evaluation during development and deployment. & \textbullet\textbullet \\
\cline{2-3}
& Evaluation should include automated and human radiologist assessments across various general language-centered and radiology-specific tasks. & \textbullet\textbullet\textbullet \\
\hline
\multirow[t]{3}{=}{Using LLMs} & Iteratively refine and use prompt techniques like few-shot prompting (providing two or more examples alongside the task instructions) and chain-of-thought prompting (instructing the model to include intermediary steps) and retrieval-augmented generation adapted to the use case. & \textbullet \\
\cline{2-3}
& Document model details (type, version, access date) and employed prompts. & \textbullet \\
\cline{2-3}
& Consider working with LLMs that can accept image inputs. These can greatly enhance potential radiology applications but come with additional technical intricacies. & \textbullet\textbullet \\
\hline
\end{tabular}
\vspace{1mm} \\
Note—Recommendations and estimation of technical difficulty are based on recent radiology and machine learning literature, and discussions with radiologists, machine learning engineers. The provided difficulty rating is the authors' attempt to estimate the complexity of the task considering and depending on individual skills, available resources, and institutional context.
\end{table}
\section{Conclusion}
In this work, we discussed technical considerations of using LLMs in radiology, and emphasized the importance of understanding their capabilities and limitations. We argue that pretraining on radiology-specific data is impractical for most labs and application-focused departments. Instead, we recommend an iterative approach: starting with optimized prompting of general-purpose models, then enriching context through in-context learning (providing examples) and retrieval-augmented generation (providing relevant input), enabling access to suitable tools, and finally fine-tuning if necessary. For fine-tuning, meticulous dataset curation is essential. We recommend using locally deployed open models to protect sensitive data while ensuring reproducibility. Ultimately, human expert evaluation and interdisciplinary teamwork between radiologists and computer scientists are the cornerstones to guide the development of safe and effective development and use of LLMs for radiology. As the field is rapidly evolving, best practices need to be periodically reassessed, since the effectiveness of these techniques may vary across different LLM types and versions. We hope this review inspires radiologists and others in the field to explore LLM technology for improving radiology practice.

\subsection*{Acknowledgements}
C.B. received research support from the Promedica Foundation, Switzerland. A.S.C. NIH grants R01 HL167974, R01HL169345, R01 AR077604, R01 EB002524, R01 AR079431, P41 EB027060; ARPA-H grants AY2AX000045 and 1AYSAX0000024-01; NIH contracts 75N92020C00008 and 75N92020C00021 and from GE Healthcare and Philips. A.S.C. and C.P.L. were supported by NIBIB MIDRC via contracts 75N92020C00008 and 75N92020C00021. C.Z. is an employee of Hugging Face, New York City, NY, USA. K.L. is an employee of NVIDIA, New York City, NY, USA. C.P.L. reports activities not related to the present article: Board of directors and shareholder, Bunkerhill Health 3/31/2019, Option holder, Whiterabbit.ai 10/01/2017, Advisor and option holder, GalileoCDS 05/01/2019, Advisor and option holder, Sirona Medical 07/06/2020, Advisor and option holder, Adra 09/17/2020, Advisor and option holder, Kheiron 10/21/2021, Paid consultant, Sixth Street 02/07/2022, Paid consultant, Gilmartin Capital 07/18/2022. Recent grant and gift support paid to C.P.L.’s institution: BunkerHill Health, Carestream, CARPL, Clairity, GE Healthcare, Google Cloud, IBM, Kheiron, Lambda, Lunit, Microsoft, Nightingale Open Science, Philips, Siemens Healthineers, Stability.ai, Subtle Medical, VinBrain, Visiana, Whiterabbit.ai, Lowenstein Foundation, Gordon and Betty Moore Foundation. A.S.C. discloses consulting services to Patient Square Capital, Elucid Bioimaging, Skope MR, Culvert Engineering, Edge Analytics, Image Analysis Group and Chondrometrics GmbH; and is shareholder in LVIS Corp., Subtle Medical and Brain Key. 

\subsection*{Contributions}
C.B. and A.S.C. conceptualized and led the project. All authors participated in discussions and helped writing the first draft of the manuscript. A.S.C. supervised the project.

\newpage
\bibliographystyle{unsrt}
\bibliography{main}

\begin{thebibliography}{100}

\bibitem{langlotzRadiologyReportGuide2015a}
Curtis~P. Langlotz.
\newblock {\em The {Radiology} {Report}: {A} {Guide} to {Thoughtful} {Communication} for {Radiologists} and {Other} {Medical} {Professionals}}.
\newblock CreateSpace Independent Publishing Platform, November 2015.
\newblock Google-Books-ID: K3b3jgEACAAJ.

\bibitem{eloundouGPTsAreGPTs2023a}
Tyna Eloundou, Sam Manning, Pamela Mishkin, and Daniel Rock.
\newblock {GPTs} are {GPTs}: {An} {Early} {Look} at the {Labor} {Market} {Impact} {Potential} of {Large} {Language} {Models}, August 2023.
\newblock arXiv:2303.10130 [cs, econ, q-fin].

\bibitem{shenRadiologyLLMDoubleEdged2023}
Yiqiu Shen, Laura Heacock, Jonathan Elias, Keith~D. Hentel, Beatriu Reig, George Shih, and Linda Moy.
\newblock {ChatGPT} and other large language models are double-edged swords.
\newblock {\em Radiology}, 307(2):e230163, 2023.
\newblock https://doi.org/10.1148/radiol.230163.

\bibitem{adams1979hitchhiker}
Douglas Adams.
\newblock {\em The {Hitchhiker}'s {Guide} to the {Galaxy}}.
\newblock Hitchhiker series. Harmony Books, 1979.

\bibitem{bhayanaChatbotsLargeLanguage2024}
Rajesh Bhayana.
\newblock Chatbots and {Large} {Language} {Models} in {Radiology}: {A} {Practical} {Primer} for {Clinical} and {Research} {Applications}.
\newblock {\em Radiology}, 310(1):e232756, January 2024.
\newblock Publisher: Radiological Society of North America.

\bibitem{bengio2000neural}
Yoshua Bengio, Réjean Ducharme, and Pascal Vincent.
\newblock A neural probabilistic language model.
\newblock {\em Advances in neural information processing systems}, 13, 2000.

\bibitem{yu2019review}
Yong Yu, Xiaosheng Si, Changhua Hu, and Jianxun Zhang.
\newblock A review of recurrent neural networks: {LSTM} cells and network architectures.
\newblock {\em Neural {C}omputation}, 31(7):1235--1270, 2019.

\bibitem{vaswaniAttentionAllYou2017}
Ashish Vaswani, Noam Shazeer, Niki Parmar, Jakob Uszkoreit, Llion Jones, Aidan~N. Gomez, Lukasz Kaiser, and Illia Polosukhin.
\newblock Attention {Is} {All} {You} {Need}, December 2017.
\newblock arXiv:1706.03762 [cs].

\bibitem{devlin2018bert}
Jacob Devlin, Ming-Wei Chang, Kenton Lee, and Kristina Toutanova.
\newblock Bert: {Pre}-training of deep bidirectional transformers for language understanding.
\newblock 2018.
\newblock arXiv:1810.04805 [cs].

\bibitem{brownLanguageModelsAre2020}
Tom~B. Brown, Benjamin Mann, Nick Ryder, Melanie Subbiah, Jared Kaplan, Prafulla Dhariwal, Arvind Neelakantan, Pranav Shyam, Girish Sastry, Amanda Askell, Sandhini Agarwal, Ariel Herbert-Voss, Gretchen Krueger, Tom Henighan, Rewon Child, Aditya Ramesh, Daniel~M. Ziegler, Jeffrey Wu, Clemens Winter, Christopher Hesse, Mark Chen, Eric Sigler, Mateusz Litwin, Scott Gray, Benjamin Chess, Jack Clark, Christopher Berner, Sam McCandlish, Alec Radford, Ilya Sutskever, and Dario Amodei.
\newblock Language {Models} are {Few}-{Shot} {Learners}, July 2020.
\newblock arXiv:2005.14165 [cs].

\bibitem{kaplanScalingLawsNeural2020}
Jared Kaplan, Sam McCandlish, Tom Henighan, Tom~B. Brown, Benjamin Chess, Rewon Child, Scott Gray, Alec Radford, Jeffrey Wu, and Dario Amodei.
\newblock Scaling {Laws} for {Neural} {Language} {Models}, January 2020.
\newblock arXiv:2001.08361 [cs, stat].

\bibitem{van-veen-etal-2023-radadapt}
Dave Van~Veen, Cara Van~Uden, Maayane Attias, Anuj Pareek, Christian Bluethgen, Malgorzata Polacin, Wah Chiu, Jean-Benoit Delbrouck, Juan Zambrano~Chaves, Curtis Langlotz, Akshay Chaudhari, and John Pauly.
\newblock {RadAdapt}: {Radiology} report summarization via lightweight domain adaptation of large language models.
\newblock In Dina Demner-fushman, Sophia Ananiadou, and Kevin Cohen, editors, {\em The 22nd workshop on biomedical natural language processing and {BioNLP} shared tasks}, pages 449--460, Toronto, Canada, July 2023. Association for Computational Linguistics.

\bibitem{yang2019xlnet}
Zhilin Yang, Zihang Dai, Yiming Yang, Jaime Carbonell, Russ~R Salakhutdinov, and Quoc~V Le.
\newblock Xlnet: {Generalized} autoregressive pretraining for language understanding.
\newblock {\em Advances in neural information processing systems}, 32, 2019.

\bibitem{ouyangTrainingLanguageModels2022}
Long Ouyang, Jeff Wu, Xu~Jiang, Diogo Almeida, Carroll~L. Wainwright, Pamela Mishkin, Chong Zhang, Sandhini Agarwal, Katarina Slama, Alex Ray, John Schulman, Jacob Hilton, Fraser Kelton, Luke Miller, Maddie Simens, Amanda Askell, Peter Welinder, Paul Christiano, Jan Leike, and Ryan Lowe.
\newblock Training language models to follow instructions with human feedback, March 2022.
\newblock arXiv:2203.02155 [cs].

\bibitem{chenVisualGPT2022}
Jun Chen, Han Guo, Kai Yi, Boyang Li, and Mohamed Elhoseiny.
\newblock {VisualGPT}: {Data}-efficient adaptation of pretrained language models for image captioning.
\newblock In {\em 2022 {IEEE}/{CVF} conference on computer vision and pattern recognition ({CVPR})}, pages 18009--18019, 2022.

\bibitem{liu2023visual}
Haotian Liu, Chunyuan Li, Qingyang Wu, and Yong~Jae Lee.
\newblock Visual instruction tuning.
\newblock In {\em Thirty-seventh conference on neural information processing systems}, 2023.

\bibitem{schickToolformerLanguageModels2023}
Timo Schick, Jane {Dwivedi-Yu}, Roberto Dess{\`i}, Roberta Raileanu, Maria Lomeli, Luke Zettlemoyer, Nicola Cancedda, and Thomas Scialom.
\newblock Toolformer: {{Language Models Can Teach Themselves}} to {{Use Tools}}, February 2023.
\newblock arXiv:2302.04761 [cs].

\bibitem{nakanoWebGPTBrowserassistedQuestionanswering2022}
Reiichiro Nakano, Jacob Hilton, Suchir Balaji, Jeff Wu, Long Ouyang, Christina Kim, Christopher Hesse, Shantanu Jain, Vineet Kosaraju, William Saunders, Xu~Jiang, Karl Cobbe, Tyna Eloundou, Gretchen Krueger, Kevin Button, Matthew Knight, Benjamin Chess, and John Schulman.
\newblock {WebGPT}: {Browser}-assisted question-answering with human feedback, June 2022.
\newblock arXiv:2112.09332 [cs].

\bibitem{bradyDevelopingPurchasingImplementing2024}
Adrian~P. Brady, Bibb Allen, Jaron Chong, Elmar Kotter, Nina Kottler, John Mongan, Lauren Oakden-Rayner, Daniel~Pinto dos Santos, An~Tang, Christoph Wald, and John Slavotinek.
\newblock Developing, {Purchasing}, {Implementing} and {Monitoring} {AI} {Tools} in {Radiology}: {Practical} {Considerations}. {A} {Multi}-{Society} {Statement} from the {ACR}, {CAR}, {ESR}, {RANZCR} and {RSNA}.
\newblock {\em Radiology: Artificial Intelligence}, 6(1):e230513, January 2024.
\newblock Publisher: Radiological Society of North America.

\bibitem{kimLargeLanguageModels2024}
Sunkyu Kim, Choong-kun Lee, and Seung-seob Kim.
\newblock Large {Language} {Models}: {A} {Guide} for {Radiologists}.
\newblock {\em Korean Journal of Radiology}, 25(2):126--133, February 2024.

\bibitem{yanStyleAwareRadiologyReport2023}
Benjamin Yan, Ruochen Liu, David~E. Kuo, Subathra Adithan, Eduardo~Pontes Reis, Stephen Kwak, Vasantha~Kumar Venugopal, Chloe~P. O'Connell, Agustina Saenz, Pranav Rajpurkar, and Michael Moor.
\newblock Style-{Aware} {Radiology} {Report} {Generation} with {RadGraph} and {Few}-{Shot} {Prompting}, October 2023.
\newblock arXiv:2310.17811 [cs].

\bibitem{schmidtGeneratingLargeLanguage2024}
Reuben~A. Schmidt, Jarrel C.~Y. Seah, Ke~Cao, Lincoln Lim, Wei Lim, and Justin Yeung.
\newblock Generating {Large} {Language} {Models} for {Detection} of {Speech} {Recognition} {Errors} in {Radiology} {Reports}.
\newblock {\em Radiology: Artificial Intelligence}, page e230205, January 2024.

\bibitem{tangEvaluatingLargeLanguage2023}
Liyan Tang, Zhaoyi Sun, Betina Idnay, Jordan~G. Nestor, Ali Soroush, Pierre~A. Elias, Ziyang Xu, Ying Ding, Greg Durrett, Justin~F. Rousseau, Chunhua Weng, and Yifan Peng.
\newblock Evaluating large language models on medical evidence summarization.
\newblock {\em npj Digital Medicine}, 6(1):1--8, August 2023.
\newblock Number: 1 Publisher: Nature Publishing Group.

\bibitem{kollerWhyWeSupport2023}
Daphne Koller, Andrew Beam, Arjun Manrai, Euan Ashley, Xiaoxuan Liu, Judy Gichoya, Chris Holmes, James Zou, Noa Dagan, Tien~Y. Wong, David Blumenthal, Isaac Kohane, and {the editors and editorial board of NEJM AI}.
\newblock Why {We} {Support} and {Encourage} the {Use} of {Large} {Language} {Models} in {NEJM} {AI} {Submissions}.
\newblock {\em NEJM AI}, 1(1):AIe2300128, December 2023.
\newblock Publisher: Massachusetts Medical Society.

\bibitem{geisEthicsArtificialIntelligence2019}
J.~Raymond Geis, Adrian~P. Brady, Carol~C. Wu, Jack Spencer, Erik Ranschaert, Jacob~L. Jaremko, Steve~G. Langer, Andrea Borondy~Kitts, Judy Birch, William~F. Shields, Robert van den Hoven~van Genderen, Elmar Kotter, Judy Wawira~Gichoya, Tessa~S. Cook, Matthew~B. Morgan, An~Tang, Nabile~M. Safdar, and Marc Kohli.
\newblock Ethics of {Artificial} {Intelligence} in {Radiology}: {Summary} of the {Joint} {European} and {North} {American} {Multisociety} {Statement}.
\newblock {\em Radiology}, 293(2):436--440, November 2019.
\newblock Publisher: Radiological Society of North America.

\bibitem{chenCheXagentFoundationModel2024b}
Zhihong Chen, Maya Varma, Jean-Benoit Delbrouck, Magdalini Paschali, Louis Blankemeier, Dave Van~Veen, Jeya Maria~Jose Valanarasu, Alaa Youssef, Joseph~Paul Cohen, Eduardo~Pontes Reis, Emily~B. Tsai, Andrew Johnston, Cameron Olsen, Tanishq~Mathew Abraham, Sergios Gatidis, Akshay~S. Chaudhari, and Curtis Langlotz.
\newblock {{CheXagent}}: {{Towards}} a {{Foundation Model}} for {{Chest X-Ray Interpretation}}, January 2024.
\newblock arXiv:2401.12208 [cs].

\bibitem{reed2022a}
Scott Reed, Konrad Zolna, Emilio Parisotto, Sergio~Gómez Colmenarejo, Alexander Novikov, Gabriel Barth-maron, Mai Giménez, Yury Sulsky, Jackie Kay, Jost~Tobias Springenberg, Tom Eccles, Jake Bruce, Ali Razavi, Ashley Edwards, Nicolas Heess, Yutian Chen, Raia Hadsell, Oriol Vinyals, Mahyar Bordbar, and Nando de~Freitas.
\newblock A generalist agent.
\newblock {\em Transactions on Machine Learning Research}, 2022.

\bibitem{gilbertLargeLanguageModel2023}
Stephen Gilbert, Hugh Harvey, Tom Melvin, Erik Vollebregt, and Paul Wicks.
\newblock Large language model {{AI}} chatbots require approval as medical devices.
\newblock {\em Nature Medicine}, pages 1--3, June 2023.

\bibitem{akincidantonoliLargeLanguageModels2023}
Tugba Akinci~D'Antonoli, Arnaldo Stanzione, Christian Bluethgen, Federica Vernuccio, Lorenzo Ugga, Michail~E. Klontzas, Renato Cuocolo, Roberto Cannella, and Burak Koçak.
\newblock Large language models in radiology: fundamentals, applications, ethical considerations, risks, and future directions.
\newblock {\em Diagnostic and Interventional Radiology (Ankara, Turkey)}, October 2023.

\bibitem{omiyeLargeLanguageModels2024}
Jesutofunmi~A. Omiye, Haiwen Gui, Shawheen~J. Rezaei, James Zou, and Roxana Daneshjou.
\newblock Large {Language} {Models} in {Medicine}: {The} {Potentials} and {Pitfalls}.
\newblock {\em Annals of Internal Medicine}, January 2024.
\newblock Publisher: American College of Physicians.

\bibitem{heSurveyLargeLanguage2023}
Kai He, Rui Mao, Qika Lin, Yucheng Ruan, Xiang Lan, Mengling Feng, and Erik Cambria.
\newblock A {{Survey}} of {{Large Language Models}} for {{Healthcare}}: From {{Data}}, {{Technology}}, and {{Applications}} to {{Accountability}} and {{Ethics}}, October 2023.

\bibitem{changSurveyEvaluationLarge2023}
Yupeng Chang, Xu~Wang, Jindong Wang, Yuan Wu, Linyi Yang, Kaijie Zhu, Hao Chen, Xiaoyuan Yi, Cunxiang Wang, Yidong Wang, Wei Ye, Yue Zhang, Yi~Chang, Philip~S. Yu, Qiang Yang, and Xing Xie.
\newblock A {Survey} on {Evaluation} of {Large} {Language} {Models}, December 2023.
\newblock arXiv:2307.03109 [cs].

\bibitem{zhaoSurveyLargeLanguage2023}
Wayne~Xin Zhao, Kun Zhou, Junyi Li, Tianyi Tang, Xiaolei Wang, Yupeng Hou, Yingqian Min, Beichen Zhang, Junjie Zhang, Zican Dong, Yifan Du, Chen Yang, Yushuo Chen, Zhipeng Chen, Jinhao Jiang, Ruiyang Ren, Yifan Li, Xinyu Tang, Zikang Liu, Peiyu Liu, Jian-Yun Nie, and Ji-Rong Wen.
\newblock A {Survey} of {Large} {Language} {Models}, November 2023.
\newblock arXiv:2303.18223 [cs].

\bibitem{smithHallucinationConfabulationNeuroanatomy2023}
Andrew~L. Smith, Felix Greaves, and Trishan Panch.
\newblock Hallucination or {Confabulation}? {Neuroanatomy} as metaphor in {Large} {Language} {Models}.
\newblock {\em PLOS Digital Health}, 2(11):e0000388, November 2023.

\bibitem{hatemChatbotConfabulationsAre2023}
Rami Hatem, Brianna Simmons, and Joseph~E. Thornton.
\newblock Chatbot {Confabulations} {Are} {Not} {Hallucinations}.
\newblock {\em JAMA Internal Medicine}, 183(10):1177, October 2023.

\bibitem{liEvaluatingObjectHallucination2023}
Yifan Li, Yifan Du, Kun Zhou, Jinpeng Wang, Xin Zhao, and Ji-Rong Wen.
\newblock Evaluating {Object} {Hallucination} in {Large} {Vision}-{Language} {Models}.
\newblock In Houda Bouamor, Juan Pino, and Kalika Bali, editors, {\em Proceedings of the 2023 {Conference} on {Empirical} {Methods} in {Natural} {Language} {Processing}}, pages 292--305, Singapore, December 2023. Association for Computational Linguistics.

\bibitem{zhaoExplainabilityLargeLanguage2024}
Haiyan Zhao, Hanjie Chen, Fan Yang, Ninghao Liu, Huiqi Deng, Hengyi Cai, Shuaiqiang Wang, Dawei Yin, and Mengnan Du.
\newblock Explainability for {Large} {Language} {Models}: {A} {Survey}.
\newblock {\em ACM Transactions on Intelligent Systems and Technology}, January 2024.
\newblock Just Accepted.

\bibitem{tonmoyComprehensiveSurveyHallucination2024}
S.~M. Towhidul~Islam Tonmoy, S.~M.~Mehedi Zaman, Vinija Jain, Anku Rani, Vipula Rawte, Aman Chadha, and Amitava Das.
\newblock A {Comprehensive} {Survey} of {Hallucination} {Mitigation} {Techniques} in {Large} {Language} {Models}, January 2024.
\newblock arXiv:2401.01313 [cs].

\bibitem{petroni2019language}
Fabio Petroni, Tim Rocktäschel, Sebastian Riedel, Patrick Lewis, Anton Bakhtin, Yuxiang Wu, and Alexander Miller.
\newblock Language models as knowledge bases?
\newblock In {\em Proceedings of the 2019 conference on empirical methods in natural language processing and the 9th international joint conference on natural language processing ({EMNLP}-{IJCNLP})}, pages 2463--2473, 2019.

\bibitem{truhnLargeLanguageModels2023a}
Daniel Truhn, Jorge~S. Reis-Filho, and Jakob~Nikolas Kather.
\newblock Large language models should be used as scientific reasoning engines, not knowledge databases.
\newblock {\em Nature Medicine}, 29(12):2983--2984, December 2023.
\newblock Publisher: Nature Publishing Group.

\bibitem{thirunavukarasuLargeLanguageModels2023}
Arun~James Thirunavukarasu, Darren Shu~Jeng Ting, Kabilan Elangovan, Laura Gutierrez, Ting~Fang Tan, and Daniel Shu~Wei Ting.
\newblock Large language models in medicine.
\newblock {\em Nature Medicine}, 29(8):1930--1940, August 2023.
\newblock Number: 8 Publisher: Nature Publishing Group.

\bibitem{grattafiori2024llama3herdmodels}
Aaron Grattafiori, Abhimanyu Dubey, Abhinav Jauhri, Abhinav Pandey, Abhishek Kadian, Ahmad Al-Dahle, Aiesha Letman, Akhil Mathur, Alan Schelten, Alex Vaughan, Amy Yang, Angela Fan, Anirudh Goyal, Anthony Hartshorn, Aobo Yang, Archi Mitra, Archie Sravankumar, Artem Korenev, Arthur Hinsvark, Arun Rao, Aston Zhang, Aurelien Rodriguez, Austen Gregerson, Ava Spataru, Baptiste Roziere, Bethany Biron, Binh Tang, Bobbie Chern, Charlotte Caucheteux, Chaya Nayak, Chloe Bi, Chris Marra, Chris McConnell, Christian Keller, Christophe Touret, Chunyang Wu, Corinne Wong, Cristian~Canton Ferrer, Cyrus Nikolaidis, Damien Allonsius, Daniel Song, Danielle Pintz, Danny Livshits, Danny Wyatt, David Esiobu, Dhruv Choudhary, Dhruv Mahajan, Diego Garcia-Olano, Diego Perino, Dieuwke Hupkes, Egor Lakomkin, Ehab AlBadawy, Elina Lobanova, Emily Dinan, Eric~Michael Smith, Filip Radenovic, Francisco Guzmán, Frank Zhang, Gabriel Synnaeve, Gabrielle Lee, Georgia~Lewis Anderson, Govind Thattai, Graeme Nail, Gregoire Mialon, Guan Pang,
  Guillem Cucurell, Hailey Nguyen, Hannah Korevaar, Hu~Xu, Hugo Touvron, Iliyan Zarov, Imanol~Arrieta Ibarra, Isabel Kloumann, Ishan Misra, Ivan Evtimov, Jack Zhang, Jade Copet, Jaewon Lee, Jan Geffert, Jana Vranes, Jason Park, Jay Mahadeokar, Jeet Shah, Jelmer van~der Linde, Jennifer Billock, Jenny Hong, Jenya Lee, Jeremy Fu, Jianfeng Chi, Jianyu Huang, Jiawen Liu, Jie Wang, Jiecao Yu, Joanna Bitton, Joe Spisak, Jongsoo Park, Joseph Rocca, Joshua Johnstun, Joshua Saxe, Junteng Jia, Kalyan~Vasuden Alwala, Karthik Prasad, Kartikeya Upasani, Kate Plawiak, Ke~Li, Kenneth Heafield, Kevin Stone, Khalid El-Arini, Krithika Iyer, Kshitiz Malik, Kuenley Chiu, Kunal Bhalla, Kushal Lakhotia, Lauren Rantala-Yeary, Laurens van~der Maaten, Lawrence Chen, Liang Tan, Liz Jenkins, Louis Martin, Lovish Madaan, Lubo Malo, Lukas Blecher, Lukas Landzaat, Luke de~Oliveira, Madeline Muzzi, Mahesh Pasupuleti, Mannat Singh, Manohar Paluri, Marcin Kardas, Maria Tsimpoukelli, Mathew Oldham, Mathieu Rita, Maya Pavlova, Melanie Kambadur,
  Mike Lewis, Min Si, Mitesh~Kumar Singh, Mona Hassan, Naman Goyal, Narjes Torabi, Nikolay Bashlykov, Nikolay Bogoychev, Niladri Chatterji, Ning Zhang, Olivier Duchenne, Onur Çelebi, Patrick Alrassy, Pengchuan Zhang, Pengwei Li, Petar Vasic, Peter Weng, Prajjwal Bhargava, Pratik Dubal, Praveen Krishnan, Punit~Singh Koura, Puxin Xu, Qing He, Qingxiao Dong, Ragavan Srinivasan, Raj Ganapathy, Ramon Calderer, Ricardo~Silveira Cabral, Robert Stojnic, Roberta Raileanu, Rohan Maheswari, Rohit Girdhar, Rohit Patel, Romain Sauvestre, Ronnie Polidoro, Roshan Sumbaly, Ross Taylor, Ruan Silva, Rui Hou, Rui Wang, Saghar Hosseini, Sahana Chennabasappa, Sanjay Singh, Sean Bell, Seohyun~Sonia Kim, Sergey Edunov, Shaoliang Nie, Sharan Narang, Sharath Raparthy, Sheng Shen, Shengye Wan, Shruti Bhosale, Shun Zhang, Simon Vandenhende, Soumya Batra, Spencer Whitman, Sten Sootla, Stephane Collot, Suchin Gururangan, Sydney Borodinsky, Tamar Herman, Tara Fowler, Tarek Sheasha, Thomas Georgiou, Thomas Scialom, Tobias Speckbacher,
  Todor Mihaylov, Tong Xiao, Ujjwal Karn, Vedanuj Goswami, Vibhor Gupta, Vignesh Ramanathan, Viktor Kerkez, Vincent Gonguet, Virginie Do, Vish Vogeti, Vítor Albiero, Vladan Petrovic, Weiwei Chu, Wenhan Xiong, Wenyin Fu, Whitney Meers, Xavier Martinet, Xiaodong Wang, Xiaofang Wang, Xiaoqing~Ellen Tan, Xide Xia, Xinfeng Xie, Xuchao Jia, Xuewei Wang, Yaelle Goldschlag, Yashesh Gaur, Yasmine Babaei, Yi~Wen, Yiwen Song, Yuchen Zhang, Yue Li, Yuning Mao, Zacharie~Delpierre Coudert, Zheng Yan, Zhengxing Chen, Zoe Papakipos, Aaditya Singh, Aayushi Srivastava, Abha Jain, Adam Kelsey, Adam Shajnfeld, Adithya Gangidi, Adolfo Victoria, Ahuva Goldstand, Ajay Menon, Ajay Sharma, Alex Boesenberg, Alexei Baevski, Allie Feinstein, Amanda Kallet, Amit Sangani, Amos Teo, Anam Yunus, Andrei Lupu, Andres Alvarado, Andrew Caples, Andrew Gu, Andrew Ho, Andrew Poulton, Andrew Ryan, Ankit Ramchandani, Annie Dong, Annie Franco, Anuj Goyal, Aparajita Saraf, Arkabandhu Chowdhury, Ashley Gabriel, Ashwin Bharambe, Assaf Eisenman, Azadeh
  Yazdan, Beau James, Ben Maurer, Benjamin Leonhardi, Bernie Huang, Beth Loyd, Beto~De Paola, Bhargavi Paranjape, Bing Liu, Bo~Wu, Boyu Ni, Braden Hancock, Bram Wasti, Brandon Spence, Brani Stojkovic, Brian Gamido, Britt Montalvo, Carl Parker, Carly Burton, Catalina Mejia, Ce~Liu, Changhan Wang, Changkyu Kim, Chao Zhou, Chester Hu, Ching-Hsiang Chu, Chris Cai, Chris Tindal, Christoph Feichtenhofer, Cynthia Gao, Damon Civin, Dana Beaty, Daniel Kreymer, Daniel Li, David Adkins, David Xu, Davide Testuggine, Delia David, Devi Parikh, Diana Liskovich, Didem Foss, Dingkang Wang, Duc Le, Dustin Holland, Edward Dowling, Eissa Jamil, Elaine Montgomery, Eleonora Presani, Emily Hahn, Emily Wood, Eric-Tuan Le, Erik Brinkman, Esteban Arcaute, Evan Dunbar, Evan Smothers, Fei Sun, Felix Kreuk, Feng Tian, Filippos Kokkinos, Firat Ozgenel, Francesco Caggioni, Frank Kanayet, Frank Seide, Gabriela~Medina Florez, Gabriella Schwarz, Gada Badeer, Georgia Swee, Gil Halpern, Grant Herman, Grigory Sizov, Guangyi, Zhang, Guna
  Lakshminarayanan, Hakan Inan, Hamid Shojanazeri, Han Zou, Hannah Wang, Hanwen Zha, Haroun Habeeb, Harrison Rudolph, Helen Suk, Henry Aspegren, Hunter Goldman, Hongyuan Zhan, Ibrahim Damlaj, Igor Molybog, Igor Tufanov, Ilias Leontiadis, Irina-Elena Veliche, Itai Gat, Jake Weissman, James Geboski, James Kohli, Janice Lam, Japhet Asher, Jean-Baptiste Gaya, Jeff Marcus, Jeff Tang, Jennifer Chan, Jenny Zhen, Jeremy Reizenstein, Jeremy Teboul, Jessica Zhong, Jian Jin, Jingyi Yang, Joe Cummings, Jon Carvill, Jon Shepard, Jonathan McPhie, Jonathan Torres, Josh Ginsburg, Junjie Wang, Kai Wu, Kam~Hou U, Karan Saxena, Kartikay Khandelwal, Katayoun Zand, Kathy Matosich, Kaushik Veeraraghavan, Kelly Michelena, Keqian Li, Kiran Jagadeesh, Kun Huang, Kunal Chawla, Kyle Huang, Lailin Chen, Lakshya Garg, Lavender A, Leandro Silva, Lee Bell, Lei Zhang, Liangpeng Guo, Licheng Yu, Liron Moshkovich, Luca Wehrstedt, Madian Khabsa, Manav Avalani, Manish Bhatt, Martynas Mankus, Matan Hasson, Matthew Lennie, Matthias Reso, Maxim
  Groshev, Maxim Naumov, Maya Lathi, Meghan Keneally, Miao Liu, Michael~L. Seltzer, Michal Valko, Michelle Restrepo, Mihir Patel, Mik Vyatskov, Mikayel Samvelyan, Mike Clark, Mike Macey, Mike Wang, Miquel~Jubert Hermoso, Mo~Metanat, Mohammad Rastegari, Munish Bansal, Nandhini Santhanam, Natascha Parks, Natasha White, Navyata Bawa, Nayan Singhal, Nick Egebo, Nicolas Usunier, Nikhil Mehta, Nikolay~Pavlovich Laptev, Ning Dong, Norman Cheng, Oleg Chernoguz, Olivia Hart, Omkar Salpekar, Ozlem Kalinli, Parkin Kent, Parth Parekh, Paul Saab, Pavan Balaji, Pedro Rittner, Philip Bontrager, Pierre Roux, Piotr Dollar, Polina Zvyagina, Prashant Ratanchandani, Pritish Yuvraj, Qian Liang, Rachad Alao, Rachel Rodriguez, Rafi Ayub, Raghotham Murthy, Raghu Nayani, Rahul Mitra, Rangaprabhu Parthasarathy, Raymond Li, Rebekkah Hogan, Robin Battey, Rocky Wang, Russ Howes, Ruty Rinott, Sachin Mehta, Sachin Siby, Sai~Jayesh Bondu, Samyak Datta, Sara Chugh, Sara Hunt, Sargun Dhillon, Sasha Sidorov, Satadru Pan, Saurabh Mahajan,
  Saurabh Verma, Seiji Yamamoto, Sharadh Ramaswamy, Shaun Lindsay, Shaun Lindsay, Sheng Feng, Shenghao Lin, Shengxin~Cindy Zha, Shishir Patil, Shiva Shankar, Shuqiang Zhang, Shuqiang Zhang, Sinong Wang, Sneha Agarwal, Soji Sajuyigbe, Soumith Chintala, Stephanie Max, Stephen Chen, Steve Kehoe, Steve Satterfield, Sudarshan Govindaprasad, Sumit Gupta, Summer Deng, Sungmin Cho, Sunny Virk, Suraj Subramanian, Sy~Choudhury, Sydney Goldman, Tal Remez, Tamar Glaser, Tamara Best, Thilo Koehler, Thomas Robinson, Tianhe Li, Tianjun Zhang, Tim Matthews, Timothy Chou, Tzook Shaked, Varun Vontimitta, Victoria Ajayi, Victoria Montanez, Vijai Mohan, Vinay~Satish Kumar, Vishal Mangla, Vlad Ionescu, Vlad Poenaru, Vlad~Tiberiu Mihailescu, Vladimir Ivanov, Wei Li, Wenchen Wang, Wenwen Jiang, Wes Bouaziz, Will Constable, Xiaocheng Tang, Xiaojian Wu, Xiaolan Wang, Xilun Wu, Xinbo Gao, Yaniv Kleinman, Yanjun Chen, Ye~Hu, Ye~Jia, Ye~Qi, Yenda Li, Yilin Zhang, Ying Zhang, Yossi Adi, Youngjin Nam, Yu, Wang, Yu~Zhao, Yuchen Hao, Yundi
  Qian, Yunlu Li, Yuzi He, Zach Rait, Zachary DeVito, Zef Rosnbrick, Zhaoduo Wen, Zhenyu Yang, Zhiwei Zhao, and Zhiyu Ma.
\newblock The {Llama} 3 {Herd} of {Models}, 2024.
\newblock arXiv:2407.21783 [cs.AI].

\bibitem{saab2024capabilitiesgeminimodelsmedicine}
Khaled Saab, Tao Tu, Wei-Hung Weng, Ryutaro Tanno, David Stutz, Ellery Wulczyn, Fan Zhang, Tim Strother, Chunjong Park, Elahe Vedadi, Juanma~Zambrano Chaves, Szu-Yeu Hu, Mike Schaekermann, Aishwarya Kamath, Yong Cheng, David G.~T. Barrett, Cathy Cheung, Basil Mustafa, Anil Palepu, Daniel McDuff, Le~Hou, Tomer Golany, Luyang Liu, Jean baptiste Alayrac, Neil Houlsby, Nenad Tomasev, Jan Freyberg, Charles Lau, Jonas Kemp, Jeremy Lai, Shekoofeh Azizi, Kimberly Kanada, SiWai Man, Kavita Kulkarni, Ruoxi Sun, Siamak Shakeri, Luheng He, Ben Caine, Albert Webson, Natasha Latysheva, Melvin Johnson, Philip Mansfield, Jian Lu, Ehud Rivlin, Jesper Anderson, Bradley Green, Renee Wong, Jonathan Krause, Jonathon Shlens, Ewa Dominowska, S.~M.~Ali Eslami, Katherine Chou, Claire Cui, Oriol Vinyals, Koray Kavukcuoglu, James Manyika, Jeff Dean, Demis Hassabis, Yossi Matias, Dale Webster, Joelle Barral, Greg Corrado, Christopher Semturs, S.~Sara Mahdavi, Juraj Gottweis, Alan Karthikesalingam, and Vivek Natarajan.
\newblock Capabilities of {Gemini} {Models} in {Medicine}, 2024.
\newblock arXiv:2404.18416 [cs.AI].

\bibitem{savageDiagnosticReasoningPrompts2024a}
Thomas Savage, Ashwin Nayak, Robert Gallo, Ekanath Rangan, and Jonathan~H. Chen.
\newblock Diagnostic reasoning prompts reveal the potential for large language model interpretability in medicine.
\newblock {\em npj Digital Medicine}, 7(1):1--7, January 2024.
\newblock Number: 1 Publisher: Nature Publishing Group.

\bibitem{gallegosBiasFairnessLarge2023}
Isabel~O. Gallegos, Ryan~A. Rossi, Joe Barrow, Md~Mehrab Tanjim, Sungchul Kim, Franck Dernoncourt, Tong Yu, Ruiyi Zhang, and Nesreen~K. Ahmed.
\newblock Bias and {Fairness} in {Large} {Language} {Models}: {A} {Survey}, September 2023.
\newblock arXiv:2309.00770 [cs].

\bibitem{thakurUnveilingGenderBias2023}
Vishesh Thakur.
\newblock Unveiling {Gender} {Bias} in {Terms} of {Profession} {Across} {LLMs}: {Analyzing} and {Addressing} {Sociological} {Implications}, August 2023.
\newblock arXiv:2307.09162 [cs].

\bibitem{koccakbias}
Burak Ko{\c{c}}ak, Andrea Ponsiglione, Arnaldo Stanzione, Christian Bluethgen, Jo{\~a}o Santinha, Lorenzo Ugga, Merel Huisman, Michail~E Klontzas, Roberto Cannella, and Renato Cuocolo.
\newblock Bias in artificial intelligence for medical imaging: fundamentals, detection, avoidance, mitigation, challenges, ethics, and prospects.
\newblock {\em Diagnostic and interventional radiology (Ankara, Turkey)}.

\bibitem{carliniExtractingTrainingData2021}
Nicholas Carlini, Florian Tram{\`e}r, Eric Wallace, Matthew Jagielski, Ariel {Herbert-Voss}, Katherine Lee, Adam Roberts, Tom Brown, Dawn Song, {\'U}lfar Erlingsson, Alina Oprea, and Colin Raffel.
\newblock Extracting {{Training Data}} from {{Large Language Models}}.
\newblock In {\em 30th {{USENIX Security Symposium}} ({{USENIX Security}} 21)}, pages 2633--2650, 2021.

\bibitem{priyanshuAreChatbotsReady2023}
Aman Priyanshu, Supriti Vijay, Ayush Kumar, Rakshit Naidu, and Fatemehsadat Mireshghallah.
\newblock Are {Chatbots} {Ready} for {Privacy}-{Sensitive} {Applications}? {An} {Investigation} into {Input} {Regurgitation} and {Prompt}-{Induced} {Sanitization}, May 2023.
\newblock arXiv:2305.15008 [cs].

\bibitem{AnnouncingOpenAIBug}
Announcing {OpenAI}’s {Bug} {Bounty} {Program}.
\newblock https://openai.com/blog/bug-bounty-program [acc. 2024-02-08].

\bibitem{bagdasaryanAbusingImagesSounds2023}
Eugene Bagdasaryan, Tsung-Yin Hsieh, Ben Nassi, and Vitaly Shmatikov.
\newblock Abusing {Images} and {Sounds} for {Indirect} {Instruction} {Injection} in {Multi}-{Modal} {LLMs}, October 2023.
\newblock arXiv:2307.10490 [cs].

\bibitem{vanveenClinicalTextSummarization2023}
Dave Van~Veen, Cara Van~Uden, Louis Blankemeier, Jean-Benoit Delbrouck, Asad Aali, Christian Bluethgen, Anuj Pareek, Malgorzata Polacin, Eduardo~Pontes Reis, Anna Seehofnerova, Nidhi Rohatgi, Poonam Hosamani, William Collins, Neera Ahuja, Curtis~P. Langlotz, Jason Hom, Sergios Gatidis, John Pauly, and Akshay~S. Chaudhari.
\newblock Clinical {Text} {Summarization}: {Adapting} {Large} {Language} {Models} {Can} {Outperform} {Human} {Experts}, October 2023.
\newblock arXiv:2309.07430 [cs].

\bibitem{anthonyCaseCoDesigningModel2024}
Quentin Anthony, Jacob Hatef, Deepak Narayanan, Stella Biderman, Stas Bekman, Junqi Yin, Aamir Shafi, Hari Subramoni, and Dhabaleswar Panda.
\newblock The {Case} for {Co}-{Designing} {Model} {Architectures} with {Hardware}, January 2024.
\newblock arXiv:2401.14489 [cs].

\bibitem{rajbhandariZeROMemoryOptimizations2020}
Samyam Rajbhandari, Jeff Rasley, Olatunji Ruwase, and Yuxiong He.
\newblock {ZeRO}: {Memory} {Optimizations} {Toward} {Training} {Trillion} {Parameter} {Models}, May 2020.
\newblock arXiv:1910.02054 [cs, stat].

\bibitem{soldainiDolmaOpenCorpus2024}
Luca Soldaini, Rodney Kinney, Akshita Bhagia, Dustin Schwenk, David Atkinson, Russell Authur, Ben Bogin, Khyathi Chandu, Jennifer Dumas, Yanai Elazar, Valentin Hofmann, Ananya~Harsh Jha, Sachin Kumar, Li~Lucy, Xinxi Lyu, Nathan Lambert, Ian Magnusson, Jacob Morrison, Niklas Muennighoff, Aakanksha Naik, Crystal Nam, Matthew~E. Peters, Abhilasha Ravichander, Kyle Richardson, Zejiang Shen, Emma Strubell, Nishant Subramani, Oyvind Tafjord, Pete Walsh, Luke Zettlemoyer, Noah~A. Smith, Hannaneh Hajishirzi, Iz~Beltagy, Dirk Groeneveld, Jesse Dodge, and Kyle Lo.
\newblock Dolma: an {Open} {Corpus} of {Three} {Trillion} {Tokens} for {Language} {Model} {Pretraining} {Research}, January 2024.
\newblock arXiv:2402.00159 [cs].

\bibitem{zhengJudgingLLMasaJudgeMTBench2023}
Lianmin Zheng, Wei-Lin Chiang, Ying Sheng, Siyuan Zhuang, Zhanghao Wu, Yonghao Zhuang, Zi~Lin, Zhuohan Li, Dacheng Li, Eric~P. Xing, Hao Zhang, Joseph~E. Gonzalez, and Ion Stoica.
\newblock Judging {LLM}-as-a-{Judge} with {MT}-{Bench} and {Chatbot} {Arena}, December 2023.
\newblock arXiv:2306.05685 [cs].

\bibitem{alpaca}
Rohan Taori, Ishaan Gulrajani, Tianyi Zhang, Yann Dubois, Xuechen Li, Carlos Guestrin, Percy Liang, and {Tatsunori B. Hashimoto}.
\newblock Stanford {Alpaca}: {An} instruction-following {LLaMA} model, 2023.
\newblock https://github.com/tatsu-lab/stanford\_alpaca [acc. 2024-01-25].

\bibitem{singhalExpertLevelMedicalQuestion2023a}
Karan Singhal, Tao Tu, Juraj Gottweis, Rory Sayres, Ellery Wulczyn, Le~Hou, Kevin Clark, Stephen Pfohl, Heather Cole-Lewis, Darlene Neal, Mike Schaekermann, Amy Wang, Mohamed Amin, Sami Lachgar, Philip Mansfield, Sushant Prakash, Bradley Green, Ewa Dominowska, Blaise Aguera~y Arcas, Nenad Tomasev, Yun Liu, Renee Wong, Christopher Semturs, S.~Sara Mahdavi, Joelle Barral, Dale Webster, Greg~S. Corrado, Yossi Matias, Shekoofeh Azizi, Alan Karthikesalingam, and Vivek Natarajan.
\newblock Towards {Expert}-{Level} {Medical} {Question} {Answering} with {Large} {Language} {Models}, May 2023.
\newblock arXiv:2305.09617 [cs].

\bibitem{jiangMistral7B2023}
Albert~Q. Jiang, Alexandre Sablayrolles, Arthur Mensch, Chris Bamford, Devendra~Singh Chaplot, Diego de~las Casas, Florian Bressand, Gianna Lengyel, Guillaume Lample, Lucile Saulnier, Lélio~Renard Lavaud, Marie-Anne Lachaux, Pierre Stock, Teven~Le Scao, Thibaut Lavril, Thomas Wang, Timothée Lacroix, and William~El Sayed.
\newblock Mistral {7B}, October 2023.
\newblock arXiv:2310.06825 [cs].

\bibitem{fuTinyTitansCan2024}
Xue-Yong Fu, Md~Tahmid~Rahman Laskar, Elena Khasanova, Cheng Chen, and Shashi~Bhushan TN.
\newblock Tiny {Titans}: {Can} {Smaller} {Large} {Language} {Models} {Punch} {Above} {Their} {Weight} in the {Real} {World} for {Meeting} {Summarization}?, February 2024.
\newblock arXiv:2402.00841 [cs].

\bibitem{jiangMixtralExperts2024}
Albert~Q. Jiang, Alexandre Sablayrolles, Antoine Roux, Arthur Mensch, Blanche Savary, Chris Bamford, Devendra~Singh Chaplot, Diego de~las Casas, Emma~Bou Hanna, Florian Bressand, Gianna Lengyel, Guillaume Bour, Guillaume Lample, Lélio~Renard Lavaud, Lucile Saulnier, Marie-Anne Lachaux, Pierre Stock, Sandeep Subramanian, Sophia Yang, Szymon Antoniak, Teven~Le Scao, Théophile Gervet, Thibaut Lavril, Thomas Wang, Timothée Lacroix, and William~El Sayed.
\newblock Mixtral of {Experts}, January 2024.
\newblock arXiv:2401.04088 [cs].

\bibitem{guMambaLinearTimeSequence2023}
Albert Gu and Tri Dao.
\newblock Mamba: {Linear}-{Time} {Sequence} {Modeling} with {Selective} {State} {Spaces}, December 2023.
\newblock arXiv:2312.00752 [cs].

\bibitem{shahCreationAdoptionLarge2023}
Nigam~H. Shah, David Entwistle, and Michael~A. Pfeffer.
\newblock Creation and {Adoption} of {Large} {Language} {Models} in {Medicine}.
\newblock {\em JAMA}, 330(9):866--869, September 2023.

\bibitem{touvronLlamaOpenFoundation2023}
Hugo Touvron, Louis Martin, Kevin Stone, Peter Albert, Amjad Almahairi, Yasmine Babaei, Nikolay Bashlykov, Soumya Batra, Prajjwal Bhargava, Shruti Bhosale, Dan Bikel, Lukas Blecher, Cristian~Canton Ferrer, Moya Chen, Guillem Cucurull, David Esiobu, Jude Fernandes, Jeremy Fu, Wenyin Fu, Brian Fuller, Cynthia Gao, Vedanuj Goswami, Naman Goyal, Anthony Hartshorn, Saghar Hosseini, Rui Hou, Hakan Inan, Marcin Kardas, Viktor Kerkez, Madian Khabsa, Isabel Kloumann, Artem Korenev, Punit~Singh Koura, Marie-Anne Lachaux, Thibaut Lavril, Jenya Lee, Diana Liskovich, Yinghai Lu, Yuning Mao, Xavier Martinet, Todor Mihaylov, Pushkar Mishra, Igor Molybog, Yixin Nie, Andrew Poulton, Jeremy Reizenstein, Rashi Rungta, Kalyan Saladi, Alan Schelten, Ruan Silva, Eric~Michael Smith, Ranjan Subramanian, Xiaoqing~Ellen Tan, Binh Tang, Ross Taylor, Adina Williams, Jian~Xiang Kuan, Puxin Xu, Zheng Yan, Iliyan Zarov, Yuchen Zhang, Angela Fan, Melanie Kambadur, Sharan Narang, Aurelien Rodriguez, Robert Stojnic, Sergey Edunov, and Thomas
  Scialom.
\newblock Llama 2: {Open} {Foundation} and {Fine}-{Tuned} {Chat} {Models}, July 2023.
\newblock arXiv:2307.09288 [cs].

\bibitem{huyenRLHFReinforcementLearning2023a}
Chip Huyen.
\newblock {RLHF}: {Reinforcement} {Learning} from {Human} {Feedback}, May 2023.

\bibitem{xiongDoctorGLMFinetuningYour2023}
Honglin Xiong, Sheng Wang, Yitao Zhu, Zihao Zhao, Yuxiao Liu, Linlin Huang, Qian Wang, and Dinggang Shen.
\newblock {DoctorGLM}: {Fine}-tuning your {Chinese} {Doctor} is not a {Herculean} {Task}, April 2023.
\newblock arXiv:2304.01097 [cs].

\bibitem{liLLaVAMedTrainingLarge2023}
Chunyuan Li, Cliff Wong, Sheng Zhang, Naoto Usuyama, Haotian Liu, Jianwei Yang, Tristan Naumann, Hoifung Poon, and Jianfeng Gao.
\newblock {LLaVA}-{Med}: {Training} a {Large} {Language}-and-{Vision} {Assistant} for {Biomedicine} in {One} {Day}, June 2023.
\newblock arXiv:2306.00890 [cs].

\bibitem{howard2018FineTunedLanguage}
Jeremy Howard and Sebastian Ruder.
\newblock Fine-tuned language models for text classification.
\newblock {\em CoRR}, abs/1801.06146, 2018.

\bibitem{hanMedAlpacaOpenSourceCollection2023}
Tianyu Han, Lisa~C. Adams, Jens-Michalis Papaioannou, Paul Grundmann, Tom Oberhauser, Alexander Löser, Daniel Truhn, and Keno~K. Bressem.
\newblock {MedAlpaca} -- {An} {Open}-{Source} {Collection} of {Medical} {Conversational} {AI} {Models} and {Training} {Data}, October 2023.
\newblock arXiv:2304.08247 [cs].

\bibitem{kahnLongTail}
Charles~E. Kahn.
\newblock The {Long} {Tail}.
\newblock {https://pubs.rsna.org/page/ai/blog/2019/4/the\_long\_tail [acc. 2024-02-08]}.

\bibitem{muennighoffCrosslingualGeneralizationMultitask2023}
Niklas Muennighoff, Thomas Wang, Lintang Sutawika, Adam Roberts, Stella Biderman, Teven~Le Scao, M.~Saiful Bari, Sheng Shen, Zheng-Xin Yong, Hailey Schoelkopf, Xiangru Tang, Dragomir Radev, Alham~Fikri Aji, Khalid Almubarak, Samuel Albanie, Zaid Alyafeai, Albert Webson, Edward Raff, and Colin Raffel.
\newblock Crosslingual {Generalization} through {Multitask} {Finetuning}, May 2023.
\newblock arXiv:2211.01786 [cs].

\bibitem{liangHolisticEvaluationLanguage2023}
Percy Liang, Rishi Bommasani, Tony Lee, Dimitris Tsipras, Dilara Soylu, Michihiro Yasunaga, Yian Zhang, Deepak Narayanan, Yuhuai Wu, Ananya Kumar, Benjamin Newman, Binhang Yuan, Bobby Yan, Ce~Zhang, Christian Cosgrove, Christopher~D. Manning, Christopher Ré, Diana Acosta-Navas, Drew~A. Hudson, Eric Zelikman, Esin Durmus, Faisal Ladhak, Frieda Rong, Hongyu Ren, Huaxiu Yao, Jue Wang, Keshav Santhanam, Laurel Orr, Lucia Zheng, Mert Yuksekgonul, Mirac Suzgun, Nathan Kim, Neel Guha, Niladri Chatterji, Omar Khattab, Peter Henderson, Qian Huang, Ryan Chi, Sang~Michael Xie, Shibani Santurkar, Surya Ganguli, Tatsunori Hashimoto, Thomas Icard, Tianyi Zhang, Vishrav Chaudhary, William Wang, Xuechen Li, Yifan Mai, Yuhui Zhang, and Yuta Koreeda.
\newblock Holistic {Evaluation} of {Language} {Models}, October 2023.
\newblock arXiv:2211.09110 [cs].

\bibitem{rafailovDirectPreferenceOptimization2023}
Rafael Rafailov, Archit Sharma, Eric Mitchell, Stefano Ermon, Christopher~D. Manning, and Chelsea Finn.
\newblock Direct {Preference} {Optimization}: {Your} {Language} {Model} is {Secretly} a {Reward} {Model}, December 2023.
\newblock arXiv:2305.18290 [cs].

\bibitem{ethayarajhKTOModelAlignment2024}
Kawin Ethayarajh, Winnie Xu, Niklas Muennighoff, Dan Jurafsky, and Douwe Kiela.
\newblock {KTO}: {Model} {Alignment} as {Prospect} {Theoretic} {Optimization}, February 2024.
\newblock arXiv:2402.01306 [cs].

\bibitem{liSelfAlignmentInstructionBacktranslation2023}
Xian Li, Ping Yu, Chunting Zhou, Timo Schick, Luke Zettlemoyer, Omer Levy, Jason Weston, and Mike Lewis.
\newblock Self-{Alignment} with {Instruction} {Backtranslation}, August 2023.
\newblock arXiv: 2308.06259v2 [cs].

\bibitem{wang-etal-2023-self-instruct}
Yizhong Wang, Yeganeh Kordi, Swaroop Mishra, Alisa Liu, Noah~A. Smith, Daniel Khashabi, and Hannaneh Hajishirzi.
\newblock Self-instruct: {Aligning} language models with self-generated instructions.
\newblock In Anna Rogers, Jordan Boyd-Graber, and Naoaki Okazaki, editors, {\em Proceedings of the 61st annual meeting of the association for computational linguistics (volume 1: {Long} papers)}, pages 13484--13508, Toronto, Canada, July 2023. Association for Computational Linguistics.

\bibitem{geroSelfVerificationImprovesFewShot2023}
Zelalem Gero, Chandan Singh, Hao Cheng, Tristan Naumann, Michel Galley, Jianfeng Gao, and Hoifung Poon.
\newblock Self-{Verification} {Improves} {Few}-{Shot} {Clinical} {Information} {Extraction}, May 2023.
\newblock arXiv: 2306.00024v1 [cs].

\bibitem{huLoRALowRankAdaptation2021}
Edward~J. Hu, Yelong Shen, Phillip Wallis, Zeyuan Allen-Zhu, Yuanzhi Li, Shean Wang, Lu~Wang, and Weizhu Chen.
\newblock {LoRA}: {Low}-{Rank} {Adaptation} of {Large} {Language} {Models}.
\newblock June 2021.
\newblock arXiv:2106.09685v2 [cs].

\bibitem{singhalLargeLanguageModels2023}
Karan Singhal, Shekoofeh Azizi, Tao Tu, S.~Sara Mahdavi, Jason Wei, Hyung~Won Chung, Nathan Scales, Ajay Tanwani, Heather Cole-Lewis, Stephen Pfohl, Perry Payne, Martin Seneviratne, Paul Gamble, Chris Kelly, Abubakr Babiker, Nathanael Schärli, Aakanksha Chowdhery, Philip Mansfield, Dina Demner-Fushman, Blaise Agüera~y Arcas, Dale Webster, Greg~S. Corrado, Yossi Matias, Katherine Chou, Juraj Gottweis, Nenad Tomasev, Yun Liu, Alvin Rajkomar, Joelle Barral, Christopher Semturs, Alan Karthikesalingam, and Vivek Natarajan.
\newblock Large language models encode clinical knowledge.
\newblock {\em Nature}, 620(7972):172--180, August 2023.
\newblock Number: 7972 Publisher: Nature Publishing Group.

\bibitem{lesterPowerScaleParameterEfficient2021}
Brian Lester, Rami Al-Rfou, and Noah Constant.
\newblock The {Power} of {Scale} for {Parameter}-{Efficient} {Prompt} {Tuning}, September 2021.
\newblock arXiv:2104.08691 [cs].

\bibitem{liuTuningLanguageModels2024}
Alisa Liu, Xiaochuang Han, Yizhong Wang, Yulia Tsvetkov, Yejin Choi, and Noah~A. Smith.
\newblock Tuning {Language} {Models} by {Proxy}, January 2024.
\newblock arXiv:2401.08565 [cs].

\bibitem{mangrulkar2022PEFT}
Sourab Mangrulkar, Sylvain Gugger, Lysandre Debut, Younes Belkada, Sayak Paul, and Benjamin Bossan.
\newblock {PEFT}: {State}-of-the-art parameter-efficient fine-tuning methods, 2022.
\newblock https://github.com/huggingface/peft [acc. 2024-02-01].

\bibitem{dettmersQLoRAEfficientFinetuning2023}
Tim Dettmers, Artidoro Pagnoni, Ari Holtzman, and Luke Zettlemoyer.
\newblock {QLoRA}: {Efficient} {Finetuning} of {Quantized} {LLMs}, May 2023.
\newblock arXiv:2305.14314 [cs].

\bibitem{wangDatasetDistillation2020}
Tongzhou Wang, Jun-Yan Zhu, Antonio Torralba, and Alexei~A. Efros.
\newblock Dataset {Distillation}, February 2020.
\newblock arXiv:1811.10959 [cs, stat].

\bibitem{hsiehDistillingStepbyStepOutperforming2023}
Cheng-Yu Hsieh, Chun-Liang Li, Chih-Kuan Yeh, Hootan Nakhost, Yasuhisa Fujii, Alexander Ratner, Ranjay Krishna, Chen-Yu Lee, and Tomas Pfister.
\newblock Distilling {Step}-by-{Step}! {Outperforming} {Larger} {Language} {Models} with {Less} {Training} {Data} and {Smaller} {Model} {Sizes}, July 2023.
\newblock arXiv:2305.02301 [cs].

\bibitem{Ollama}
Ollama.
\newblock https://ollama.com [acc. 2024-02-01].

\bibitem{huangMultimodalFoundationModels2024}
Shih-Cheng Huang, Malte Jensen, Serena {Yeung-Levy}, Matthew~P. Lungren, Hoifung Poon, and Akshay~S. Chaudhari.
\newblock Multimodal {{Foundation Models}} for {{Medical Imaging}} - {{A Systematic Review}} and {{Implementation Guidelines}}, October 2024.
\newblock medRxiv:2024.10.23.24316003.

\bibitem{zhangMMLLMsRecentAdvances2024}
Duzhen Zhang, Yahan Yu, Chenxing Li, Jiahua Dong, Dan Su, Chenhui Chu, and Dong Yu.
\newblock {MM}-{LLMs}: {Recent} {Advances} in {MultiModal} {Large} {Language} {Models}, January 2024.
\newblock arXiv:2401.13601 [cs].

\bibitem{noriCanGeneralistFoundation2023}
Harsha Nori, Yin~Tat Lee, Sheng Zhang, Dean Carignan, Richard Edgar, Nicolo Fusi, Nicholas King, Jonathan Larson, Yuanzhi Li, Weishung Liu, Renqian Luo, Scott~Mayer McKinney, Robert~Osazuwa Ness, Hoifung Poon, Tao Qin, Naoto Usuyama, Chris White, and Eric Horvitz.
\newblock Can {Generalist} {Foundation} {Models} {Outcompete} {Special}-{Purpose} {Tuning}? {Case} {Study} in {Medicine}, November 2023.
\newblock arXiv:2311.16452 [cs].

\bibitem{yangDawnLMMsPreliminary2023}
Zhengyuan Yang, Linjie Li, Kevin Lin, Jianfeng Wang, Chung-Ching Lin, Zicheng Liu, and Lijuan Wang.
\newblock The {Dawn} of {LMMs}: {Preliminary} {Explorations} with {GPT}-{4V}(ision), October 2023.
\newblock arXiv:2309.17421 [cs].

\bibitem{tuGeneralistBiomedicalAI2023}
Tao Tu, Shekoofeh Azizi, Danny Driess, Mike Schaekermann, Mohamed Amin, Pi-Chuan Chang, Andrew Carroll, Chuck Lau, Ryutaro Tanno, Ira Ktena, Basil Mustafa, Aakanksha Chowdhery, Yun Liu, Simon Kornblith, David Fleet, Philip Mansfield, Sushant Prakash, Renee Wong, Sunny Virmani, Christopher Semturs, S.~Sara Mahdavi, Bradley Green, Ewa Dominowska, Blaise Aguera~y Arcas, Joelle Barral, Dale Webster, Greg~S. Corrado, Yossi Matias, Karan Singhal, Pete Florence, Alan Karthikesalingam, and Vivek Natarajan.
\newblock Towards {Generalist} {Biomedical} {AI}, July 2023.
\newblock arXiv:2307.14334 [cs].

\bibitem{wuGeneralistFoundationModel2023}
Chaoyi Wu, Xiaoman Zhang, Ya~Zhang, Yanfeng Wang, and Weidi Xie.
\newblock Towards {Generalist} {Foundation} {Model} for {Radiology} by {Leveraging} {Web}-scale {2D}\&{3D} {Medical} {Data}, November 2023.
\newblock arXiv:2308.02463 [cs].

\bibitem{pellegriniRaDialogLargeVisionLanguage2023}
Chantal Pellegrini, Ege Özsoy, Benjamin Busam, Nassir Navab, and Matthias Keicher.
\newblock {RaDialog}: {A} {Large} {Vision}-{Language} {Model} for {Radiology} {Report} {Generation} and {Conversational} {Assistance}, November 2023.
\newblock arXiv:2311.18681 [cs].

\bibitem{leeCXRLLAVAMultimodalLarge2024}
Seowoo Lee, Jiwon Youn, Hyungjin Kim, Mansu Kim, and Soon~Ho Yoon.
\newblock {CXR}-{LLAVA}: a multimodal large language model for interpreting chest {X}-ray images, January 2024.
\newblock arXiv:2310.18341 [cs].

\bibitem{bidermanPythiaSuiteAnalyzing2023}
Stella Biderman, Hailey Schoelkopf, Quentin Anthony, Herbie Bradley, Kyle O'Brien, Eric Hallahan, Mohammad~Aflah Khan, Shivanshu Purohit, USVSN~Sai Prashanth, Edward Raff, Aviya Skowron, Lintang Sutawika, and Oskar van~der Wal.
\newblock Pythia: {A} {Suite} for {Analyzing} {Large} {Language} {Models} {Across} {Training} and {Scaling}, May 2023.
\newblock arXiv:2304.01373 [cs].

\bibitem{goodmanAIGeneratedClinicalSummaries2024}
Katherine~E. Goodman, Paul~H. Yi, and Daniel~J. Morgan.
\newblock {AI}-{Generated} {Clinical} {Summaries} {Require} {More} {Than} {Accuracy}.
\newblock {\em JAMA}, January 2024.

\bibitem{papineni2002bleu}
Kishore Papineni, Salim Roukos, Todd Ward, and Wei-Jing Zhu.
\newblock {BLEU}: A method for automatic evaluation of machine translation.
\newblock In {\em Proceedings of the 40th Annual Meeting of the {{Association}} for {{Computational Linguistics}}}, pages 311--318, 2002.

\bibitem{lin2004rouge}
Chin-Yew Lin.
\newblock Rouge: {{A}} package for automatic evaluation of summaries.
\newblock In {\em Text Summarization Branches Out}, pages 74--81, 2004.

\bibitem{yimACIBENCHNovelAmbient2023}
Wen-wai Yim, Yujuan Fu, Asma~Ben Abacha, Neal Snider, Thomas Lin, and Meliha Yetisgen.
\newblock {ACI}-{BENCH}: a {Novel} {Ambient} {Clinical} {Intelligence} {Dataset} for {Benchmarking} {Automatic} {Visit} {Note} {Generation}, June 2023.

\bibitem{zhang2019bertscore}
Tianyi Zhang, Varsha Kishore, Felix Wu, Kilian~Q Weinberger, and Yoav Artzi.
\newblock {{BERTScore}}: {{Evaluating}} text generation with {{BERT}}.
\newblock In {\em International Conference on Learning Representations}, 2019.

\bibitem{zhangOptimizingFactualCorrectness2020}
Yuhao Zhang, Derek Merck, Emily Tsai, Christopher~D. Manning, and Curtis Langlotz.
\newblock Optimizing the {Factual} {Correctness} of a {Summary}: {A} {Study} of {Summarizing} {Radiology} {Reports}.
\newblock In Dan Jurafsky, Joyce Chai, Natalie Schluter, and Joel Tetreault, editors, {\em Proceedings of the 58th {Annual} {Meeting} of the {Association} for {Computational} {Linguistics}}, pages 5108--5120, Online, July 2020. Association for Computational Linguistics.

\bibitem{delbrouckImprovingFactualCorrectness2022a}
Jean-Benoit Delbrouck, Pierre Chambon, Christian Bluethgen, Emily Tsai, Omar Almusa, and Curtis Langlotz.
\newblock Improving the {Factual} {Correctness} of {Radiology} {Report} {Generation} with {Semantic} {Rewards}.
\newblock In Yoav Goldberg, Zornitsa Kozareva, and Yue Zhang, editors, {\em Findings of the {Association} for {Computational} {Linguistics}: {EMNLP} 2022}, pages 4348--4360, Abu Dhabi, United Arab Emirates, December 2022. Association for Computational Linguistics.

\bibitem{jin2019pubmedqa}
Qiao Jin, Bhuwan Dhingra, Zhengping Liu, William Cohen, and Xinghua Lu.
\newblock {PubMedQA}: {A} dataset for biomedical research question answering.
\newblock In {\em Proceedings of the 2019 conference on empirical methods in natural language processing and the 9th international joint conference on natural language processing ({EMNLP}-{IJCNLP})}, pages 2567--2577, 2019.

\bibitem{hendryckstest2021}
Dan Hendrycks, Collin Burns, Steven Basart, Andy Zou, Mantas Mazeika, Dawn Song, and Jacob Steinhardt.
\newblock Measuring massive multitask language understanding.
\newblock {\em Proceedings of the International Conference on Learning Representations (ICLR)}, 2021.

\bibitem{wang2018glue}
Alex Wang, Amanpreet Singh, Julian Michael, Felix Hill, Omer Levy, and Samuel~R. Bowman.
\newblock {GLUE}: {A} multi-task benchmark and analysis platform for natural language understanding.
\newblock In {\em International conference on learning representations}, 2019.

\bibitem{chavesRaLEsBenchmarkRadiology2023}
Juan Manuel~Zambrano Chaves, Nandita Bhaskhar, Maayane Attias, Jean-Benoit Delbrouck, Daniel Rubin, Andreas~Markus Loening, Curtis Langlotz, and Akshay~S. Chaudhari.
\newblock {RaLEs}: a {Benchmark} for {Radiology} {Language} {Evaluations}.
\newblock In {\em Thirty-seventh Conference on Neural Information Processing Systems Datasets and Benchmarks Track}, November 2023.

\bibitem{zakkaAlmanacRetrievalAugmentedLanguage2024}
Cyril Zakka, Rohan Shad, Akash Chaurasia, Alex~R. Dalal, Jennifer~L. Kim, Michael Moor, Robyn Fong, Curran Phillips, Kevin Alexander, Euan Ashley, Jack Boyd, Kathleen Boyd, Karen Hirsch, Curt Langlotz, Rita Lee, Joanna Melia, Joanna Nelson, Karim Sallam, Stacey Tullis, Melissa~Ann Vogelsong, John~Patrick Cunningham, and William Hiesinger.
\newblock Almanac — {Retrieval}-{Augmented} {Language} {Models} for {Clinical} {Medicine}.
\newblock {\em NEJM AI}, 1(2):AIoa2300068, January 2024.
\newblock Publisher: Massachusetts Medical Society.

\bibitem{truhnEcologicalFootprintMedical2023}
Daniel Truhn, Gustav Müller-Franzes, and Jakob~Nikolas Kather.
\newblock The ecological footprint of medical {AI}.
\newblock {\em European Radiology}, August 2023.

\bibitem{pattersonCarbonEmissionsLarge}
David Patterson, Joseph Gonzalez, Quoc Le, Chen Liang, Lluis-Miquel Munguia, Daniel Rothchild, David So, Maud Texier, and Jeff Dean.
\newblock Carbon emissions and large neural network training.
\newblock {\em arXiv}, 2021.
\newblock arXiv:2104.10350 [cs].

\bibitem{dooEvaluationClimateAwareMetrics2023}
Florence~X. Doo, Vishwa~S. Parekh, Adway Kanhere, Dharmam Savani, Ali~S. Tejani, Amir Sapkota, and Paul~H. Yi.
\newblock Evaluation of {Climate}-{Aware} {Metrics} {Tools} for {Radiology} {Informatics} and {Artificial} {Intelligence}: {Toward} a {Potential} {Radiology} {Ecolabel}.
\newblock {\em Journal of the American College of Radiology}, December 2023.

\bibitem{zamfirescu-pereiraWhyJohnnyCant2023}
J.D. {Zamfirescu-Pereira}, Richmond~Y. Wong, Bjoern Hartmann, and Qian Yang.
\newblock Why {{Johnny Can}}'t {{Prompt}}: {{How Non-AI Experts Try}} (and {{Fail}}) to {{Design LLM Prompts}}.
\newblock In {\em Proceedings of the 2023 {{CHI Conference}} on {{Human Factors}} in {{Computing Systems}}}, {{CHI}} '23, pages 1--21, New York, NY, USA, April 2023. Association for Computing Machinery.

\bibitem{weng2023prompt}
Lilian Weng.
\newblock Prompt engineering.
\newblock {\em lilianweng.github.io}, March 2023.
\newblock https://lilianweng.github.io/posts/2023-03-15-prompt-engineering/ [acc. 2024-02-01].

\bibitem{OpenAIPlatform}
{OpenAI} {Platform}.
\newblock https://platform.openai.com [acc. 2024-02-07].

\bibitem{PromptEngineeringGuide}
Prompt {Engineering} {Guide}.
\newblock https://www.promptingguide.ai/ [acc. 2024-02-07].

\bibitem{adamsLeveragingGPT4Post2023}
Lisa~C. Adams, Daniel Truhn, Felix Busch, Avan Kader, Stefan~M. Niehues, Marcus~R. Makowski, and Keno~K. Bressem.
\newblock Leveraging {GPT}-4 for {Post} {Hoc} {Transformation} of {Free}-text {Radiology} {Reports} into {Structured} {Reporting}: {A} {Multilingual} {Feasibility} {Study}.
\newblock {\em Radiology}, 307(4):e230725, May 2023.
\newblock Publisher: Radiological Society of North America.

\bibitem{samuel_colvin_2024_10619320}
Samuel Colvin, David Montague, Adrian~Garcia Badaracco, Hasan Ramezani, Eric Jolibois, Marcelo Trylesinski, Sydney Runkle, Terrence Dorsey, Serge Matveenko, pyup.io bot, David Hewitt, Arseny Boykov, Sebastián Ramírez, {Viicos}, Nikita Grishko, Koudai Aono, Alex Hall, Yurii Karabas, Vitaly Samigullin, Stephen~Brown II, Yasser Tahiri, Amin Alaee, Davis Kirkendall, {layday}, Daniel Smith, Marc Mueller, Nuno André, and {retnikt}.
\newblock Pydantic, February 2024.
\newblock https://doi.org/10.5281/zenodo.8180180.

\bibitem{holtzmanCuriousCaseNeural2020}
Ari Holtzman, Jan Buys, Li~Du, Maxwell Forbes, and Yejin Choi.
\newblock The {Curious} {Case} of {Neural} {Text} {Degeneration}, February 2020.
\newblock arXiv:1904.09751 [cs].

\bibitem{lewisRetrievalAugmentedGenerationKnowledgeIntensive2021}
Patrick Lewis, Ethan Perez, Aleksandra Piktus, Fabio Petroni, Vladimir Karpukhin, Naman Goyal, Heinrich Küttler, Mike Lewis, Wen-tau Yih, Tim Rocktäschel, Sebastian Riedel, and Douwe Kiela.
\newblock Retrieval-{Augmented} {Generation} for {Knowledge}-{Intensive} {NLP} {Tasks}, April 2021.
\newblock arXiv:2005.11401 [cs].

\bibitem{jinMedCPTContrastivePretrained2023}
Qiao Jin, Won Kim, Qingyu Chen, Donald~C Comeau, Lana Yeganova, W~John Wilbur, and Zhiyong Lu.
\newblock {MedCPT}: {Contrastive} {Pre}-trained {Transformers} with large-scale {PubMed} search logs for zero-shot biomedical information retrieval.
\newblock {\em Bioinformatics}, 39(11):btad651, November 2023.

\bibitem{ovadiaFineTuningRetrievalComparing2024}
Oded Ovadia, Menachem Brief, Moshik Mishaeli, and Oren Elisha.
\newblock Fine-{Tuning} or {Retrieval}? {Comparing} {Knowledge} {Injection} in {LLMs}, January 2024.
\newblock arXiv:2312.05934 [cs].

\end{thebibliography}

\newpage
% \appendix
% \input{sections/suppl}

%%%%%%%%%%%%%%%%%%%%%%%%%%%%%%%%%%%%%%%%%%%%%%%%%%%%%%%%%%%%

\end{document}